\pdfoutput=1

\documentclass[11pt]{article}

\usepackage[final]{acl}

\usepackage{float}
\usepackage{times}
\usepackage{latexsym}
\usepackage{comment}
\usepackage{booktabs}
\usepackage{arydshln}
\usepackage{xcolor}
\usepackage{spverbatim}
\usepackage{fancyvrb}
\usepackage{multirow}
\usepackage[normalem]{ulem}
\useunder{\uline}{\ul}{}
\usepackage{amsmath}
\usepackage{caption}

\makeatletter
\def\VerbLB{\FV@Command{}{VerbLB}}
\begingroup
\catcode`\^^M=\active%
\gdef\FVC@VerbLB#1{%
  \begingroup%
    \FV@UseKeyValues%
    \FV@FormattingPrep%
    \FV@CatCodes%
    \def^^M{ }%
    \catcode`#1=12%
    \def\@tempa{\def\FancyVerbGetVerb####1####2}%
    \expandafter\@tempa\string#1{\mbox{##2}\endgroup}%
    \FancyVerbGetVerb\FV@EOL}%
\endgroup
\makeatother

\usepackage[T1]{fontenc} 

\usepackage[utf8]{inputenc}

\usepackage{microtype}

\usepackage{inconsolata}

\usepackage{graphicx}

\title{Zero-Shot Keyphrase Generation: Investigating Specialized Instructions and Multi-Sample Aggregation on Large Language Models}

\author{Jayanth Mohan\thanks{Both authors contributed equally to this research.} \mbox{    }\mbox{    } Jishnu Ray Chowdhury$^{*}$\thanks{Most work done at the University of Illinois Chicago unaffiliated to the author's current position at Bloomberg.}\mbox{    }\mbox{    } Tomas Malik \mbox{    }\mbox{    } Cornelia Caragea \\
  Computer Science \\
  University of Illinois Chicago \\
  \texttt{jmoha11@uic.edu\mbox{    }\mbox{    } jraych2@uic.edu\mbox{    }\mbox{    } tmalik6@uic.edu\mbox{    }\mbox{    } cornelia@uic.edu} \\
  \\}

\begin{document}
\maketitle
\begin{abstract}
Keyphrases are the essential topical phrases that summarize a document. Keyphrase generation is a long-standing NLP task for automatically generating keyphrases for a given document. While the task has been comprehensively explored in the past via various models, only a few works perform some preliminary analysis of Large Language Models (LLMs) for the task. Given the impact of LLMs in the field of NLP, it is important to conduct a more thorough examination of their potential for keyphrase generation. In this paper, we attempt to meet this demand with our research agenda. Specifically, we focus on the zero-shot capabilities of open-source instruction-tuned LLMs (Phi-3, Llama-3) and the closed-source GPT-4o for this task. We systematically investigate the effect of providing task-relevant specialized instructions in the prompt. Moreover, we design task-specific counterparts to self-consistency-style strategies for LLMs and show significant benefits from our proposals over the baselines.
\end{abstract}

\section{Introduction}

Keyphrases are concise, representative phrases that encapsulate the most essential and relevant topical information in a document \cite{HasanN14}. They serve as a high-level summary, providing quick insight into the text. Keyphrases can be ``present'' if they appear verbatim in the text, or ``absent'' if they are semantically implied and do not occur explicitly in the text. While keyphrase extraction focuses on identifying present keyphrases \cite{ParkC23,PatelC21,AlzaidyCG19, bennani-smires-etal-2018-simple, YuN18, FlorescuC17, SterckxCDD16,GollapalliC14}, keyphrase generation (KPG) extends the task to include both present and absent keyphrases \cite{GargCC23,ChowdhuryPKC22,garg-etal-2022-keyphrase,meng-etal-2017-deep,yuan2020one, chan-etal-2019-neural, chen-etal-2020-exclusive}. 
Recent advancements in keyphrase research, including this work, focus primarily on KPG, as it provides a more comprehensive summary of the document's information. Keyphrases are vital in various information retrieval and NLP applications, such as document indexing and retrieval \cite{jones1999phrasier, boudin2020keyphrase}, summarization \cite{wang2013domain,abu-jbara-radev-2011-coherent}, content recommendation \cite{augenstein-etal-2017-semeval}, and search engine optimization \cite{10.1145/1141753.1141800}. 

Various previous approaches have attempted to tackle KPG. Most of them are sequence-to-sequence approaches that are trained from scratch specifically for KPG \cite{meng-etal-2017-deep, yuan2020one, chan-etal-2019-neural, chen-etal-2020-exclusive, ye-etal-2021-one2set, thomas-vajjala-2024-improving}. More recently, some approaches explore finetuning of pre-trained language models such as BART or T5 for KPG \cite{wu-etal-2021-unikeyphrase, mayank-learning, wu-etal-2023-rethinking-model, wu-etal-2024-leveraging, choi-etal-2023-simckp}. However, the field of Natural Language Processing (NLP), on the other hand, is moving away from such approaches and towards the utilization of Large Language Models (LLMs) \cite{iyer2022opt,touvron2023llamaopenefficientfoundation} that typically have much higher parameters and are pre-trained on larger scale datasets. 
As such, naturally, there is a question as to how well such models can be operated towards KPG. A few prior works conduct some studies to answer this question, primarily investigating ChatGPT as a zero-shot generator. However, they are only preliminary studies that investigate a few variants of prompts \cite{song2023chatgpt, song2023large, martinez2023chatgpt}. Our work aims to extend such studies further. Specifically, in this paper, we aim to answer three research questions (RQ1, RQ2, RQ3) as defined below.
\vspace{1mm}

\noindent \textbf{RQ1:} \textit{Can LLMs be guided to focus specifically on present or absent keyphrases via prompts?}

\noindent As discussed before, KPG typically involves the generation of two distinct types of keyphrases---present and absent which may require distinct strategies. In \citet{song2023chatgpt}, we also find that the same prompt is not necessarily good at both present and absent generation simultaneously. Thus, the question arises if we can create separate ``specialist'' prompts - one specializing in present keyphrase generation and another specializing in absent keyphrase generation. If this succeeds, we can come up with a way to combine the specialists' results to improve both present and absent keyphrase generation performance. We describe our designed specialist prompts in $\S$\ref{sec:specialists} and show their corresponding evaluation in $\S$\ref{sec:specialist_results}.

\vspace{1mm}

\noindent \textbf{RQ2:} \textit{Do more specific instructions about controlling the number of keyphrases and/or the order of generation help LLMs?}

\noindent In our baselines, we provide basic instructions regarding formatting to enable parsing of keyphrases through downstream post-processing methods. However, there is a potential to explore the application of more detailed instructions to the models. For example, we might want to specify how we want the keyphrases to be ordered - such as most relevant keyphrases being generated before less relevant ones. Metrics such as (F$_1$@5) used in keyphrase generation, focus on some first $k$ keyphrases, so it is important for the LLMs to generate the best keyphrases first. We might also want to instruct the model more specifically to not overgenerate. We find that LLMs tend to generate more keyphrases on average compared to other smaller models, which can lower precision. 
We design specific instructions corresponding to these points in $\S$\ref{sec:instruction_modifiers} and experimentally investigate them in $\S$\ref{sec:instruction_modifiers_results}. 

\vspace{1mm}

\noindent \textbf{RQ3:} \textit{Can multiple samplings of an LLM from the same input prompt be leveraged to improve performance in keyphrase generation?}

\noindent Often in KPG, beam search is used to create multiple sequences of keyphrases to improve the recall of keyphrases \cite{ChowdhuryPKC22,thomas-vajjala-2024-improving,yuan2020one}. On the other hand, the use of multiple samplings has been successful with LLMs in general NLP tasks as well. For instance, the self-consistency strategy leverages majority voting (or other aggregation techniques) across multiple sampled results for a question to boost the performance of LLMs \cite{wang2023selfconsistency}. Given the success of self-consistency (on general NLP tasks) and beam search (for KPG), we raise the question if we can similarly leverage multiple sampling from LLMs for KPG specifically. To answer this question, we devise various multi-sampling aggregation strategies for KPG in $\S$\ref{sec:multisampling} and demonstrate their corresponding results experimentally in $\S$\ref{sec:multisampling_results}.

\vspace{1mm}

We focus primarily on open-source, instruction-tuned models, specifically LLama-3 and Phi-3, in a zero-shot setting. Additionally, we include experiments with GPT-4o to benchmark against a bigger closed-source model. In our experiments, we find that specialist prompts for our models do not help (answering RQ1 negatively) and that additional detailed instructions do not help consistently (answering RQ2 negatively). However, we find that multi-sampling can be successfully leveraged to substantially boost the performance of LLMs for KPG (answering RQ3 affirmatively).

\section{Method}
We explore the performance of two open-source instruction-tuned LLMs  - \textbf{Llama-3.0 8B Instruct} \cite{dubey2024llama} and \textbf{Phi-3.0 3.8B Mini 128K Instruct} \cite{abdin2024phi} and one closed-source LLM - \textbf{GPT-4o} (version: gpt-4o-2024-11-20) \cite{achiam2023gpt} on KPG for five different datasets in a zero-shot setting. We explain our main approaches in the following subsections. 



    

\begin{figure}[t]
\centering
\includegraphics[width=0.5\textwidth]{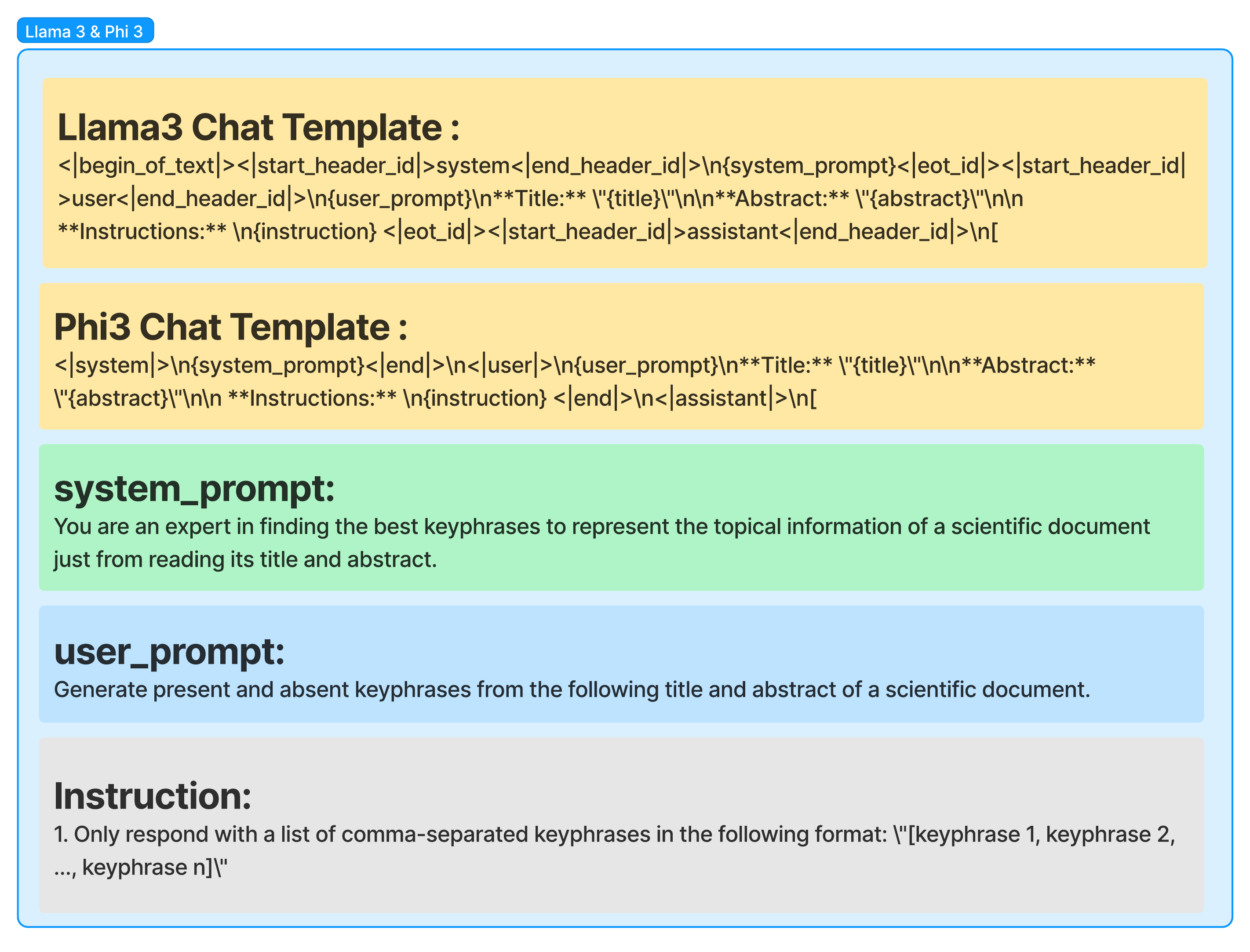}
\caption{Baseline template used for Llama-3 and Phi-3.}
\label{fig:template}
\end{figure}


\subsection{Baseline}
Here, we explain the construction of baseline prompts for Llama-3, Phi-3 and GPT-4o. First, we keep their prompt templates consistent with their corresponding chat templates as shown in Figure \ref{fig:template}. Note that at the end of the prompt, we leave an open parenthesis ``['' so that the models can directly start generating the keyphrases without any in-between irrelevant text. As can be seen in the chat templates, there are five variables: 1) the {\color{cyan!50!black}\verb+system_prompt+}, 2) the {\color{cyan!50!black}\verb+user_prompt+}, 3) the {\color{cyan!50!black}\verb+instruction+}, 4) the {\color{cyan!50!black}\verb+title+}, and 5) the {\color{cyan!50!black}\verb+abstract+}. The last two are inputs from the dataset, whereas the first three are manually defined. We define them the same way for Llama-3 and Phi-3. For GPT-4o, we use the chat completion API for sending the system prompt and user prompt. We skipped the open parenthesis for the assistant role in GPT-4o because the provided chat completion API does not support partial conversational turns. 
Our definitions for \verb+system_prompt+, \verb+user_prompt+, and \verb+instruction+ variables are also shown in Figure \ref{fig:template}. When evaluating the models on KP-Times, we changed any occurrence of ``scientific document'' with ``news article''. The user prompt is roughly inspired from the TP4 prompt template in \citet{song2023chatgpt}\footnote{Similar to \citet{song2023chatgpt}, we also verified that LLama-3 and Phi-3 can distinguish the meaning of present and absent keyphrases by themselves. Thus, we did not present any further overt definition of them in the prompts.}. We use TP4 because it presents a reasonable balance in their paper. In the baseline, the \verb+instruction+ merely provides some formatting specifications to make parsing of the keyphrases lists easier. 

\subsection{Specialist Prompts (RQ1)}
\label{sec:specialists}
As discussed before, the same prompt may not be the best for both present and absent keyphrase generation. As such, we consider if we can improve present performance and absent performance separately with ``specialist'' prompts - one dedicated to present keyphrase extraction and another to absent keyphrase generation. We design the present specialist prompt by simply changing the baseline \verb+user_prompt+ to {\color{cyan!50!black}{\texttt{``Extract present keyphrases from the following title and abstract of a scientific document.''}}} Similarly, we design the absent specialist prompt by simply changing the baseline \verb+user_prompt+ to {\color{cyan!50!black}{\texttt{``Generate absent keyphrases from the following title and abstract of a scientific document.''}}} This results in the creation of two separate prompts which we test separately. 

 \begin{figure}[t]
\centering
\includegraphics[width=0.4\textwidth]{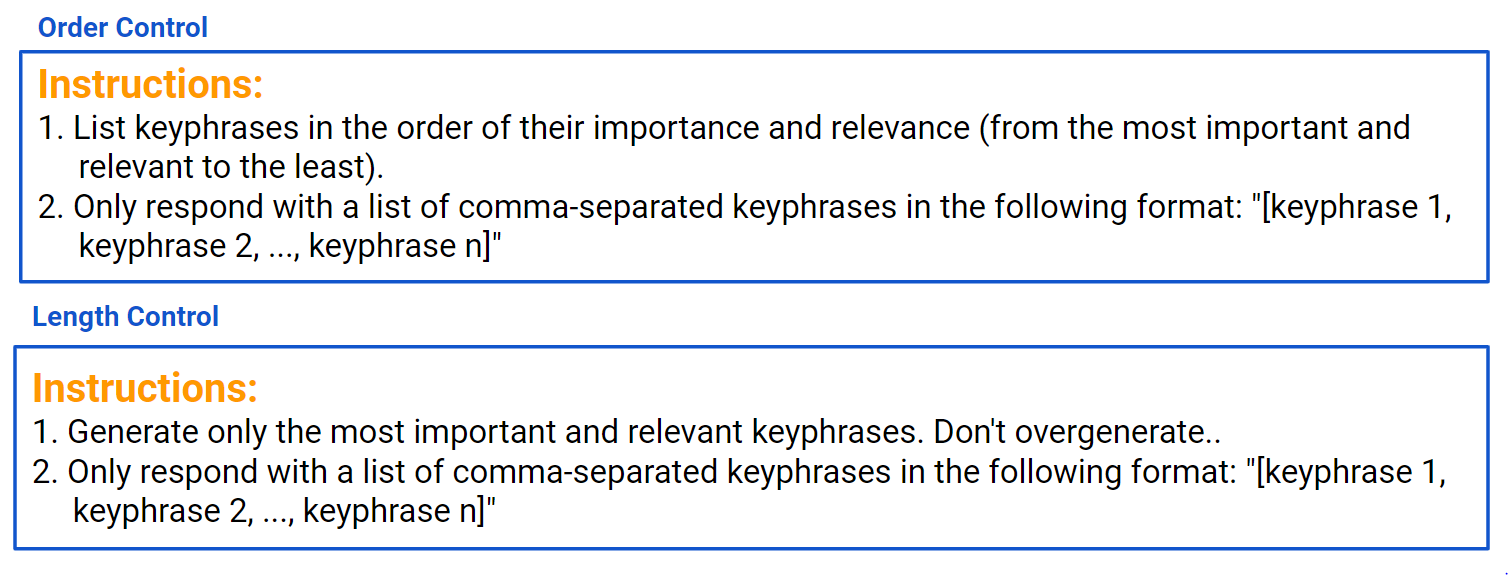}
\caption{Instructions used for Order Control and Length Control. Note that the main values for the instruction variable are in the blue bordered box. The differences of box sizes and colours are for visualization only and do not play any role in the actual prompt.}
\label{fig:instruction_mod}
\vspace{-3mm}
\end{figure}

\begin{figure*}[t]
\centering
\includegraphics[width=0.8\textwidth]{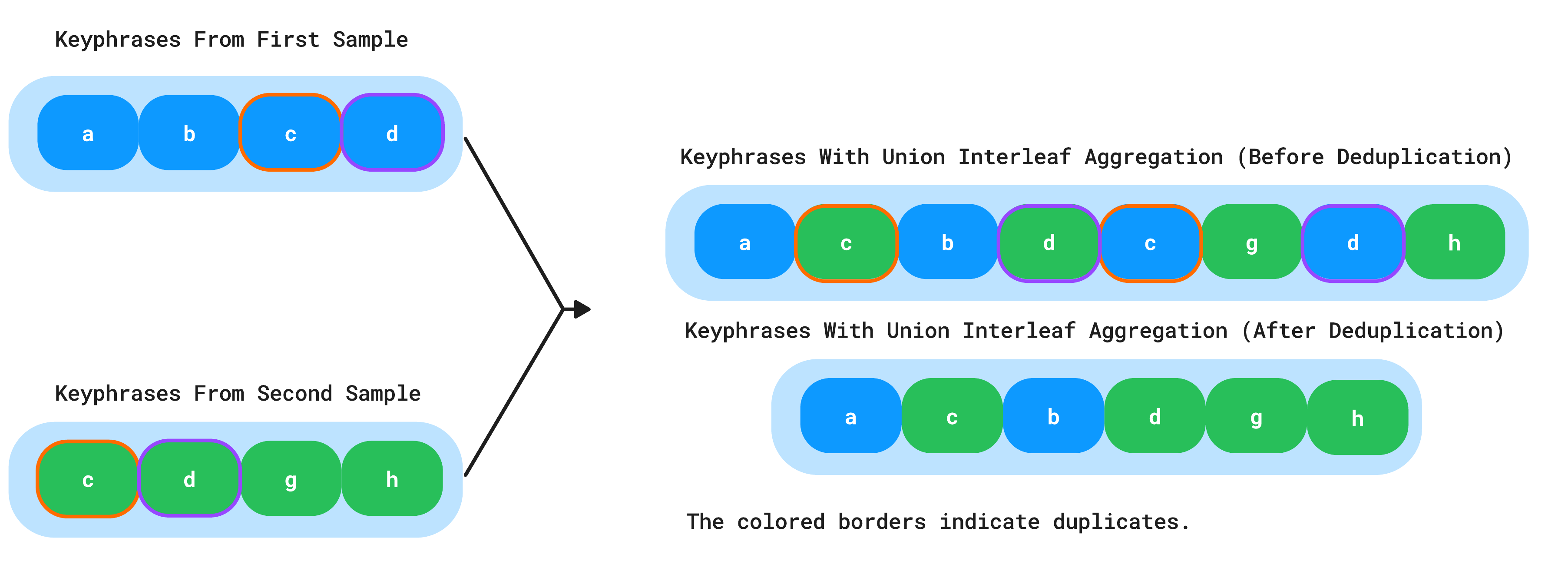}
\caption{Visualization of Union Interleaf aggregation over multiple samples.} 
\label{fig:interleaf}
\vspace{-3mm}
\end{figure*}
\subsection{Additional Instructions (RQ2)}
\label{sec:instruction_modifiers}
We consider here whether LLMs can benefit from more specific instructions as to how to order keyphrases and how many keyphrases to generate. We consider two types of instructions:
\vspace{1mm}

\noindent \textbf{1. Order Control Instruction:}  As we discussed before, the order of the keyphrases can be relevant, especially for metrics like F$_1$@5 where only the first $5$ keyphrases are kept, and we are interested in keeping the best keyphrases within the first few. So we experiment with an additional instruction that explicitly specifies the model to order the keyphrases in the descending order of relevance and importance. Concretely, we do this by changing the value of the \verb+instruction+ variable from the baseline into a numbered list having both the formatting instruction and the order control instruction as shown in Figure \ref{fig:instruction_mod}.

\noindent \textbf{2. Length Control Instruction:}  Here we focus on reducing overgeneration, which can negatively impact metrics like precision. For this, we instruct the model to generate only the most relevant keyphrases, avoiding unnecessary additions. Concretely, we do this, similar to above, by changing the \verb+instruction+ variable from the baseline as shown in Figure \ref{fig:instruction_mod}.
\vspace{1mm}

\noindent \textbf{3. Combined Control:} Here we integrate both the Order Control and Length Control instructions to prime the model to generate a concise list of keyphrases ordered by relevance. We do this by adding both the Order Control and Length Control instructions into the numbered list of instructions similar to before. In our implementation, the Length Control instruction is the first instruction (number 1), the Order Control instruction is the second one (number 2), and the formatting instruction from the baseline is the third one (number 3). 

\begin{table*}[t]
\centering
\scriptsize
\def\arraystretch{1.2}
\resizebox{1\textwidth}{!}{%
\begin{tabular}{l|cc|cc|cc|cc|cc}
\hline
 & \multicolumn{2}{c}{\textbf{Inspec}} & \multicolumn{2}{|c}{\textbf{Krapivin}} & \multicolumn{2}{|c}{\textbf{SemEval}} & \multicolumn{2}{|c}{\textbf{KP20K}} & \multicolumn{2}{|c}{\textbf{KPTimes}} \\  
\multicolumn{1}{c|}{\textbf{Models}} & \textbf{F1@M} & {\textbf{F1@5}}  & \textbf{F1@M} & {\textbf{F1@5}} & \textbf{F1@M} & {\textbf{F1@5}} & \textbf{F1@M} & {\textbf{F1@5}} & \textbf{F1@M} & \textbf{F1@5} \\ \hline
\multicolumn{11}{c}{\textbf{Present Keyphrase Generation}}\\ \hline
\multicolumn{11}{l}{\textbf{Llama-3.0 8B Instruct}}\\ \hline
Baseline & \textbf{48.3} & \textbf{40.5} & 30.9 & 32.4 & \textbf{35.5} & \textbf{36.2} & 27.7 & \textbf{30.7} & \textbf{27.0} & \textbf{31.3}\\
Present Specialist & 46.9 & 40.2 & 30.6 & 31.5 & 34.8 & 33.6 & \textbf{29.0} & 30.4 & 24.0 & 29.3\\
Absent Specialist & 47.9 & \textbf{40.5} & \textbf{31.6} & \textbf{32.8} & 35.4 & 36.0 & 28.2 & \textbf{30.7} & 22.6 & 29.5\\
\hline
\multicolumn{11}{l}{\textbf{Phi-3.0 3.8B Mini 128K Instruct}}\\ \hline
Baseline & 48.2 & 42.2 & 22.2 & 22.5 & \textbf{28.4} & \textbf{28.6} & 17.6 & \textbf{19.1} & \textbf{9.3} & \textbf{11.2}\\
Present Specialist & \textbf{48.4} & \textbf{42.6} & \textbf{22.6} & 22.6 & 26.3 & 26.0 & 17.6 & 19.0 & 8.7 & 10.5\\
Absent Specialist & 46.6 & 41.2 & \textbf{23.5} & \textbf{22.9} & 27.8 & 28.5 & \textbf{18.2} & \textbf{19.1} & 8.1 & 9.0\\
\hline
\multicolumn{11}{l}{\textbf{GPT-4o}}\\ \hline
Baseline & 56.8 & 49.7 & 26.0 & \textbf{28.0} & 33.2 & 34.1 & 20.1 & 24.7 & 11.4 & 14.7\\
Present Specialist  & \textbf{57.5} & \textbf{50.1} & \textbf{27.1} & \textbf{28.0} & \textbf{33.6} & \textbf{34.3} & \textbf{20.6} & \textbf{24.3} & \textbf{11.9} & \textbf{15.7} \\
Absent Specialist & 37.6 & 33.8 & 23.3 & 22.9 & 24.8 & 24.9 & 15.9 & 17.0 & 7.5 & 7.9\\

\hline
\multicolumn{11}{c}{\textbf{Absent Keyphrase Generation}}\\ \hline
\multicolumn{11}{l}{\textbf{Llama-3.0 8B Instruct}}\\ \hline
Baseline & \textbf{6.8} & \textbf{5.5} & \textbf{4.6} & \textbf{3.8} & \textbf{3.2} & \textbf{3.0} & 3.8 & 3.0 & \textbf{4.6} & \textbf{3.6}\\
Present Specialist & 5.6 & 4.5 & 3.4 & 2.6 & 2.5 & 2.1 & 3.4 & 2.7 & 4.1 & 3.3\\
Absent Specialist & 6.4 & 5.0 & 4.2 & \textbf{3.8} & 3.1 & \textbf{3.0} & \textbf{4.0} & \textbf{3.2} & 4.2 & \textbf{3.6}\\

\hline
\multicolumn{11}{l}{\textbf{Phi-3.0 3.8B Mini 128K Instruct}}\\ \hline
Baseline & \textbf{7.3} & \textbf{6.3} & \textbf{1.3} & 1.1 & \textbf{2.0} & \textbf{1.5} & 1.3 & 1.1 & \textbf{0.4} & \textbf{0.4}\\
Present Specialist & 7.0 & 5.6 & 1.2 & \textbf{1.2} & 1.6 & 1.3 & 1.3 & 1.1 & \textbf{0.4} & \textbf{0.4}\\
Absent Specialist & 6.6 & 5.5 & \textbf{1.3} & \textbf{1.2} & 1.7 & 1.4 & \textbf{1.4} & \textbf{1.2} & 0.3 & 0.4\\

\hline
\multicolumn{11}{l}{\textbf{GPT-4o}}\\ \hline
Baseline & 10.6 & 10.6 & \textbf{4.0} & \textbf{4.0} & 2.6 & 2.6 & 2.4 & 2.5 & 0.4 & 0.5\\
Present Specialist & \textbf{12.4} &\textbf{ 12.4} & 3.0 & 3.0 & 2.8 & 2.5 & 2.5 & 2.5 & \textbf{0.8} & \textbf{0.8}\\
Absent Specialist & 6.5 & 6.5 & 3.5 & 3.5 & \textbf{5.0} & \textbf{4.4} & \textbf{3.2} &\textbf{ 3.4} & 0.5 & 0.6\\

\hline
\end{tabular}}
\caption{Comparison of baseline prompts and specialist prompts for present and absent keyphrase generation.} 
\vspace{-4mm}
\label{tb:specialists}
\end{table*}
\begin{table*}[t]
\centering
\scriptsize
\def\arraystretch{1.2}
\begin{tabular}{l|cc|cc|cc|cc|cc}
\hline
 & \multicolumn{2}{c}{\textbf{Inspec}} & \multicolumn{2}{|c}{\textbf{Krapivin}} & \multicolumn{2}{|c}{\textbf{SemEval}} & \multicolumn{2}{|c}{\textbf{KP20K}} & \multicolumn{2}{|c}{\textbf{KPTimes}} \\  
\multicolumn{1}{c|}{\textbf{Models}} & \textbf{F1@M} & {\textbf{F1@5}}  & \textbf{F1@M} & {\textbf{F1@5}} & \textbf{F1@M} & {\textbf{F1@5}} & \textbf{F1@M} & {\textbf{F1@5}} & \textbf{F1@M} & \textbf{F1@5} \\ \hline
\multicolumn{11}{c}{\textbf{Present Keyphrase Generation}}\\ \hline
\multicolumn{11}{l}{\textbf{Llama-3.0 8B Instruct}}\\ \hline
Baseline & \textbf{48.3} & \textbf{40.5} & 30.9 & 32.4 & 35.5 & 36.2 & 27.7 & 30.7 & \textbf{27.0} & 31.3\\
Order Control & 46.0 & 38.8 & 31.1 & 33.2 & 35.8 & 36.6 & 29.0 &\textbf{ 32.1} & 24.9 & \textbf{31.5}\\
Length Control & 45.1 & 39.4 & 33.4 & 32.4 & \textbf{39.0} & \textbf{37.6} & \textbf{31.1} & 31.1 & 26.8 & 29.9\\
Combined Control & 44.4 & 38.8 & \textbf{33.5} & \textbf{33.5} & 36.8 & 36.7 & 30.9 & 31.5 & 27.1& 30.9 \\
\hline
\multicolumn{11}{l}{\textbf{Phi-3.0 3.8B Mini 128K Instruct}}\\ \hline
Baseline & \textbf{48.2} & 42.2 & 22.2 & \textbf{22.5} & \textbf{28.4} & \textbf{28.6} & 17.6 & 19.1 & \textbf{9.3} & \textbf{11.2}\\
Order Control & 45.0 & 39.5 & 21.6 & 20.8 & 25.7 & 23.6 & 16.4 & 17.7 & 7.4 & 8.0\\
Length Control &  47.8 & \textbf{42.5} & \textbf{22.8} & \textbf{22.5} & 27.5 & 26.8 & \textbf{18.2} & \textbf{19.2} & 9.0 & 10.5\\
Combined Control & 44.5 & 38.8 & 21.5 & 20.8 & 26.5 & 25.3 & 17.0 & 18.0 & 7.9 & 8.7  \\
\hline

\multicolumn{11}{l}{\textbf{GPT-4o}}\\ \hline

Baseline & \textbf{56.8} & \textbf{49.7} & 26.0 & 28.0 & \textbf{33.2} & \textbf{34.1} & 20.1 & 24.7 & 11.4 & \textbf{14.7}\\
Order Control & 54.5 & 48.3 & 24.2 & 26.5 & 30.0 & 31.8 & 18.5 & 23.4 & 9.6 & 11.9\\
Length Control  & 55.2 & 49.5 & \textbf{28.6} & \textbf{29.3} & 31.6 & 32.6 &\textbf{ 22.4} &\textbf{ 25.4} & \textbf{11.6} & 13.1\\
Combined Control & 53.3 & 47.7 & 25.8 & 27.1 & 32.7 & 33.5 & 20.7 & 23.9 & 9.8 & 10.8 \\
\hline

\multicolumn{11}{c}{\textbf{Absent Keyphrase Generation}}\\ \hline
\multicolumn{11}{l}{\textbf{Llama-3.0 8B Instruct}}\\ \hline
Baseline & \textbf{6.8} & \textbf{5.5} & 4.6 & \textbf{3.8} & \textbf{3.2} & \textbf{3.0} & 3.8 & 3.0 & \textbf{4.6} & 3.6\\
Order Control & 5.4 & 4.4 & 4.0 & 3.3 & 3.0 & 2.7 & 3.9 & \textbf{3.2} & 4.5 & \textbf{3.8}\\
Length Control & 5.3 & 4.1 & 4.2 & 3.4 & 2.4 & 2.2 & 3.9 & 3.0 & 4.4 & 3.6\\
Combined Control & 4.7 & 3.6 & \textbf{4.7} & \textbf{3.8} & 2.7 & 2.4 & \textbf{4.0} & 3.1 & 4.4 & 3.6 \\
\hline
\multicolumn{11}{l}{\textbf{Phi-3.0 3.8B Mini 128K Instruct}}\\ \hline
Baseline & \textbf{7.3} & \textbf{6.3} & 1.3 & 1.1 & \textbf{2.0} & 1.5 & \textbf{1.3} & \textbf{1.1} & \textbf{0.4} & \textbf{0.4}\\
Order Control & 6.2 & 5.2 & 1.5 & \textbf{1.3} & 1.3 & 1.2 & 1.2 & 1.0 & 0.3 & 0.3\\
Length Control & 6.8 & 5.6 & 1.5 & 1.1 & 1.9 & \textbf{1.8} & \textbf{1.3} & \textbf{1.1} & \textbf{0.4} & \textbf{0.4}\\
Combined Control & 6.4 & 5.4 & \textbf{1.6} & \textbf{1.3} & 1.2 & 1.1 & 1.2 & 1.0 & 0.3 & 0.3\\
\hline

\multicolumn{11}{l}{\textbf{GPT-4o}}\\ \hline
Baseline & 10.6 & 10.6 & \textbf{4.0} & \textbf{4.0} & \textbf{2.6} & \textbf{2.6} & \textbf{2.4} & \textbf{2.5} & 0.4 & 0.5\\
Order Control  & 9.8 & 9.5 & 2.6 & 2.6 & 2.3 & 2.2 & 2.0 & 2.1 & 0.4 & 0.5\\
Length Control & 9.9 & 9.8 & 2.2 & 2.0 & 2.5 & 2.5 & \textbf{2.4} & \textbf{2.5} & \textbf{0.6} & \textbf{0.6}\\
Combined Control & \textbf{11.0} & \textbf{11.0} & 2.5 & 2.5 & 1.8 & 1.8 & 1.9 & 1.9 & 0.3 & 0.3 \\
\hline

\end{tabular}
\caption{Comparison of baseline prompts and prompts with additional instructions for present and absent keyphrase generation.} 
\label{tb:instruction_modifier}
\end{table*}

\begin{table*}[t]
\centering
\scriptsize
\def\arraystretch{1.2}
\resizebox{0.8\textwidth}{!}{%
\begin{tabular}{l|cc|cc|cc|cc|cc}
\hline
 & \multicolumn{2}{c}{\textbf{Inspec}} & \multicolumn{2}{|c}{\textbf{Krapivin}} & \multicolumn{2}{|c}{\textbf{SemEval}} & \multicolumn{2}{|c}{\textbf{KP20K}} & \multicolumn{2}{|c}{\textbf{KPTimes}} \\  
\multicolumn{1}{c|}{\textbf{Models}} & \textbf{F1@M} & {\textbf{F1@5}}  & \textbf{F1@M} & {\textbf{F1@5}} & \textbf{F1@M} & {\textbf{F1@5}} & \textbf{F1@M} & {\textbf{F1@5}} & \textbf{F1@M} & \textbf{F1@5} \\ \hline
\multicolumn{11}{c}{\textbf{Present Keyphrase Generation}}\\ \hline
\multicolumn{11}{l}{\textbf{Llama-3.0 8B Instruct}}\\ \hline
Baseline & 48.3 & 40.5 & 30.9 & 32.4 & 35.5 & 36.2 & 27.7 & 30.7 & \textbf{27.0} & 31.3\\
\hdashline
\multicolumn{11}{l}{Multi-sampling (n=10)} \\
\hdashline
Union & 36.5 & 30.1 & 22.2 & 18.0 & 26.6 & 21.7 & 18.9 & 16.0 & 13.0 & 9.5\\
Union Concat & \textbf{50.0} & 42.6 & 30.6 & 32.2 & 37.6 & 35.1 & 27.3 & 31.0 & 23.4 & 31.4\\
Union Interleaf & 42.4 & 36.4 & 30.5 & 32.3 & \textbf{38.3} & 36.3 & \textbf{29.1} & 31.5 & 25.9 & \textbf{32.3}\\
Frequency Order & 49.9 & \textbf{45.6} & \textbf{31.8} & \textbf{33.6} & 38.0 & \textbf{38.1} & 28.7 & \textbf{32.1} & 24.7 & 31.5\\
\hline
\multicolumn{11}{l}{\textbf{Phi-3.0 3.8B Mini 128K Instruct}}\\ \hline
Baseline & 48.2 & 42.2 & 22.2 & 22.5 & 28.4 & 28.6 & 17.6 & 19.1 & 9.3 & 11.2\\
\hdashline
\multicolumn{11}{l}{Multi-sampling (n=10)} \\
\hdashline
Union & 33.8 & 29.8 & 16.9 & 15.6 & 18.7 & 14.5 & 12.9 & 11.1 & 6.7 & 5.6\\
Union Concat & 50.2 & 45.5 & 23.1 & 22.7 & 30.4 & 30.3 & 18.0 & 19.8 & 10.9 & 12.2\\
Union Interleaf & 45.2 & 41.0 & 24.9 & 25.3 & \textbf{33.2} & \textbf{31.4} & \textbf{21.6} & \textbf{22.5} & \textbf{15.1} & \textbf{14.9}\\
Frequency Order & \textbf{54.7} & \textbf{50.9} & \textbf{25.1} & \textbf{24.7} & 32.9 & 30.5 & 19.7 & 20.4 & 12.0 & 11.9\\
\hline

\multicolumn{11}{l}{\textbf{GPT-4o}}\\ \hline
Baseline & 56.8 & 49.7 & 26.0 & 28.0 & 33.2 & 34.1 & 20.1 & 24.7 & 11.4 & 14.7\\
\hdashline
\multicolumn{11}{l}{Multi-sampling (n=10)} \\
\hdashline
Union & 46.0 & 36.4 & 21.7 & 18.0 & 24.3 & 17.8 & 15.7 & 13.2 & 8.8 & 7.4 \\
Union Concat & 57.6 & 50.4 & 25.9 & 28.1 & \textbf{33.7} & 34.4 & 20.1 & 24.6 & 12.9 & 15.5\\
Union Interleaf  & 54.2 & 47.0 & \textbf{27.7} & \textbf{29.4} & 33.2 & \textbf{34.5} & \textbf{21.9} & \textbf{26.3} & \textbf{16.0} & \textbf{18.2}\\
Frequency Order & \textbf{58.2} & \textbf{52.8} & 25.9 & 25.6 & 32.7 & 31.4 & 20.0 & 21.7 & 12.1 & 12.4\\
\hline

\multicolumn{11}{c}{\textbf{Absent Keyphrase Generation}}\\ \hline
\multicolumn{11}{l}{\textbf{Llama-3.0 8B Instruct}}\\ \hline
Baseline & 6.8 & 5.5 & 4.6 & 3.8 & 3.2 & 3.0 & 3.8 & 3.0 & 4.6 & 3.6\\
\hdashline
\multicolumn{11}{l}{Multi-sampling (n=10)} \\
\hdashline
Union & 3.7 & 4.9 & 3.9 & 3.8 & 1.7 & 1.2 & 2.8 & 3.2 & 2.1 & 2.3\\
Union Concat & \textbf{8.9} & \textbf{8.2} & 5.4 & 4.7 & 3.6 & \textbf{3.6} & 4.6 & 4.5 & 4.6 & 4.4\\
Union Interleaf & 6.8 & 6.3 & 5.2 & 4.9 & 3.0 & 2.7 & 5.0 & 4.7 & 4.9 & 4.7\\
Frequency Order & 8.5 & 7.6 & \textbf{5.9} & \textbf{5.1} & \textbf{3.9} & \textbf{3.6} & \textbf{5.4} & \textbf{4.9} & \textbf{5.3} & \textbf{5.0}\\
\hline
\multicolumn{11}{l}{\textbf{Phi-3.0 3.8B Mini 128K Instruct}}\\ \hline
Baseline & 7.3 & 6.3 & 1.3 & 1.1 & 2.0 & 1.5 & 1.3 & 1.1 & 0.4 & 0.4\\
\hdashline
\multicolumn{11}{l}{Multi-sampling (n=10)} \\
\hdashline
Union &  2.7 & 2.7 & 0.8 & 0.7 & 1.1 & 1.1 & 0.8 & 0.8 & 0.2 & 0.2\\
Union Concat & 8.2 & 7.8 & 1.8 & 1.8 & 1.9 & 1.8 & 1.5 & 1.5 & 0.4 & 0.5\\
Union Interleaf & 6.2 & 6.0 & 1.7 & 1.8 & 1.3 & 0.9 & 1.6 & 1.6 & 0.4 & 0.4\\
Frequency Order & \textbf{9.3} & \textbf{9.0} & \textbf{2.1} & \textbf{2.1} & \textbf{2.5} & \textbf{2.7} & \textbf{2.0} & \textbf{2.0} & \textbf{0.6} & \textbf{0.7}\\
\hline
\multicolumn{11}{l}{\textbf{GPT-4o}}\\ \hline
Baseline & 10.6 & 10.6 & \textbf{4.0} & \textbf{4.0} & 2.6 & 2.6 & 2.4 & 2.5 & 0.4 & 0.5\\
\hdashline
\multicolumn{11}{l}{Multi-sampling (n=10)} \\
\hdashline
Union   & 5.6 & 6.8 & 1.8 & 1.9 & 1.6 & 2.1 & 1.3 & 1.5 & 0.3 & 0.2\\
Union Concat  & 10.2 & 9.7 & \textbf{3.9} & \textbf{3.7} & 2.9 & 3.3 & 2.5 & 2.3 & 0.4 & 0.5\\
Union Interleaf & 10.9 & 10.1 & 3.3 & 3.0 & 1.9 & 2.4 & \textbf{2.8} & \textbf{2.8 }& \textbf{0.6} & \textbf{0.7}\\
Frequency Order  & \textbf{11.7} & \textbf{10.8} & 3.5 & 3.1 & \textbf{3.7} & \textbf{3.6} & 2.6 & 2.6 & \textbf{0.6} & 0.6\\
\hline
\end{tabular}}
\caption{Comparison of baseline models and multisample models with different aggregation strategies for both present and absent keyphrase generation.} 
\vspace{-3.5mm}
\label{tb:multisampling}
\end{table*}

\subsection{Multi-Sampling (RQ3)}
\label{sec:multisampling}
To investigate RQ3, we stochastically generate multiple samples from LLMs with the baseline prompt using different temperatures for diversity. We take independent samples similar to self-consistency strategies \cite{wang2023selfconsistency}. In addition, in Appendix \ref{sec:beam_search}, we show that beam search, which is often used in KPG, does not improve performance on KPG, and at the same time can become expensive with large LLMs and tends to have worse diversity.  


In our multi-sampling context, for a specific input, we initially end up having a list of samples as an answer: $S = (S_1, S_2, S_3, \dots, S_n)$. Here $n$ is the number of samples. Each sample $S_i$ is a sequence of keyphrases: $S_i = (k^i_1,k^i_2,k^i_3,\dots,k^i_m)$. Each keyphrase ($k^i_j$) is a string. We describe our pipeline for processing such samples below.

\vspace{2mm}
\noindent \textbf{Ranking Samples:} We first sort the generated samples before applying any aggregation strategy in the ascending order of their perplexity. We do this because some of our aggregation techniques (e.g., Union Concatenation that we discuss below) is biased towards putting the keyphrases of the earlier samples in $S$ earlier. As we discussed before, the order of the keyphrases can be relevant for metrics like F$_1$@5. Thus, we sort them to keep the ``best'' samples according to perplexity at the forefront for any downstream aggregation. 

\vspace{2mm}
\noindent \textbf{Keyphrase Normalization:} Before aggregation, we also normalize the keyphrases using standard techniques - such as lower-casing and stemming. These are standard normalization strategies also used for evaluation to determine which keyphrases are identical. We also deduplicate each sample while preserving the order.

\vspace{2mm}
\noindent \textbf{Keyphrase Aggregation Strategies:} After ranking and normalization is done, the question is how to aggregate the results. 
We devise several strategies for aggregating the results from different samples that we discuss below. 

\vspace{2mm}
\noindent {1. \em Union: } This is a simple strategy, where we treat all the generated lists of keyphrases ($S_i$) as sets and apply union operation. The result is $\cup_{i=1}^n S_i$. All order information is destroyed in this process. 

\vspace{2mm}
\noindent {2. \em Union Concatenation: } In the context of KPG, a typical method used during beam-search to aggregate the results from multiple beams is to concatenate each of the beam sequences together (starting from the highest-ranked beam to the lowest). We simulate the same strategy here with Union Concatenation. In this approach, we concatenate all the samples: $||_{i=1}^n S_i$ (here $||$ denotes concatenation operator). After that, we deduplicate the concatenated sequence in an order-preserving manner (the first occurrence of a duplicate is the one that remains).  

\vspace{2mm}
\noindent {3. \em Union Interleaf: } In this strategy, we initially combine the samples in an interleaving pattern. That is, first we take all the first keyphrases from each sample, then all the second keyphrases from each sample, and so on. We add them to a combined list in that order. The combined list will look like: $(k^1_1, k^2_1, \dots, k^n_1, k^1_2, k^2_2, \dots k^n_2, k^1_m \dots, k^n_m)$. After this, we perform an order-preserving deduplication as in Union Concatenation. The visualization of this process is provided in Figure \ref{fig:interleaf}.  




\vspace{2mm}
\noindent {4. \em Frequency Order: } Frequency Order is the closest counterpart to majority voting as applicable for KPG. In this method, we consider the frequency of occurrence for each normalized keyphrase across all the samples. Then we sort the keyphrases in descending order of their frequency of occurrence. Thus, the highest ``voted'' (most frequent) keyphrase gets to be at the forefront of the aggregated list getting the maximum preference. In case of ties, we follow the order in Union Interleaf. That is, if there is a tie in terms of frequency between $k_1$ and $k_2$, then $k_1$ should come ahead of $k_2$ if and only if it occurs before $k_2$ in the union interleaf result for the same samples. 

\vspace{2mm}
\noindent \textbf{Dynamic Keyphrase Number Selection:} Once the aggregation is done, there is a separate question as to how to dynamically select an appropriate number of keyphrases for each input. Normally, in the baseline single sample setting, we can simply use all the keyphrases predicted by the model until the end of sequence marker. However, with increasing number of samples being aggregated, the total keyphrases can become arbitrarily high. This can lead to the overgeneration of noisy keyphrases - leading to degraded precision and F$_1$, especially for @M metrics (which considers all keyphrases by the model not just some top $k$). To resolve this, we devise an automatic protocol to dynamically select a variable number of present keyphrases and a variable number of absent keyphrases from the total generation. Concretely, we first calculate the average number of present keyphrases (say $M_{pre}$) and average number of absent keyphrases (say $M_{abs}$) per sample for a specific input.\footnote{In case the average is not a whole number, we take the ceiling.} Then from the aggregated list, we take the first $M_{pre}$ present keyphrases and the first $M_{abs}$ absent keyphrases. We treat this as the final model prediction for F$_1$@M metric calculation.

\noindent \textbf{Discussion:} A problem with Union Concatenation is that it can lead to ignoring later samples altogether due to truncating the concatenation based on either top-5 selections (for F$_1$@5) or top $M_{pre}$ and $M_{abs}$ selections (for F$_1$@M). It can be still a reasonable strategy if the concatenation is ordered such that the first few samples are of higher quality, but even with our perplexity-based ranking, it is unlikely to have that much of a difference in quality among the samples given that they are each sampled independently based on the same process. Moreover, it can be the case, that earlier keyphrases from later samples are of higher quality than later keyphrases of earlier samples. This can happen if LLMs generate the most relevant keyphrases first. Union Concat would not respect this factor. Union Interleaf or Frequency Order based aggregations, on the other hand, can address some of these points much better in theory - resulting in a better intermingling of different samples in the final list. 

\section{Experiments and Results}
For our experiments, we choose a temperature of $0.8$ which we use consistently\footnote{We chose $0.8$ because it is in the standard range of temperature typically used for self-consistency for diverse multi-sampling. We also did not find substantial differences from different temperatures in a subset of the KP20K validation set.} across all models and datasets. We explain our evaluation in Appendix \ref{sec:evaluation}.

\vspace{-2mm}
\subsection{Datasets}
In our experiments we explored a number of datasets that focus on the domain of scientific publications (SemEval \cite{kim-etal-2010-semeval}, Krapivin \cite{Krapivin2009LargeDF}, KP20K \cite{DBLP:journals/corr/abs-2004-10462}, Inspec \cite{joshi2023unsupervisedkeyphraseextractioninterpretable}), and also a dataset focusing on the news domain (KPTimes \cite{gallina2019kptimes}). These datasets are commonly used as benchmarks KPG. All experiments were performed in a zero-shot setting solely on the test subsets of the datasets. For SemEval, Krapivin, and Inspec, we utilized the full datasets across all our models: Llama-3, Phi-3, and GPT-4o. For KP20K and KPTimes, we employed the full datasets for Llama-3 and Phi-3, as they are open-source models. However, for the closed-source model GPT-4o, we used a subset of 2,000 samples from each dataset to make the experiments cost-effective. We also show the comparison between LLama-3, Phi-3, and GPT-4o on the same 2,000 samples for KPTimes and KP20K in the Appendix Table \ref{tb:multisampling_2000} and observe similar patterns as in the main paper.

\subsection{Specialist Prompts Results (RQ1)}
\label{sec:specialist_results}
In Table \ref{tb:specialists}, we present the results of our baseline and specialist (present and absent) prompts. Interestingly, we find that the specialist present and absent prompt do not consistently outperform the baseline; rather in many cases underperform compared to the baseline both in present and absent keyphrase generation. Interestingly, GPT-4o despite being estimatedly a much larger model still shows no consistent benefit from the specialized prompts; moreover, it also seems to perform worse than LLama-3 for KPG on most datasets. Thus, at least for the explored LLM-based models and the considered prompts, the answer to RQ1 seems to be negative.\footnote{As would be expected given that the specialists individually do not outperform the baseline, in our experiments, the ensembling of the two specialist models also failed to outperform the ensembling of two baseline prompt-based models.} 

\subsection{Additional Instruction Results (RQ2)}
\label{sec:instruction_modifiers_results}
In Table \ref{tb:instruction_modifier}, we present the results of including additional instructions to the baseline prompt for order control and length control as discussed before. Here, we find that Length Control can sometimes help in the performance of present keyphrase extraction in some datasets. However, the overall result is mixed, and none of the strategies of additional instructions consistently improve the baseline across both present and absent keyphrase generation. As such, the answer to RQ2 also seems to lead towards a negative outcome. 

\begin{table*}[t]
\centering
\scriptsize
\def\arraystretch{1.2}
\begin{tabular}{l|cc|cc|cc|cc|cc}
\hline
 & \multicolumn{2}{c}{\textbf{Inspec}} & \multicolumn{2}{|c}{\textbf{Krapivin}} & \multicolumn{2}{|c}{\textbf{SemEval}} & \multicolumn{2}{|c}{\textbf{KP20K}} & \multicolumn{2}{|c}{\textbf{KPTimes}} \\  
\multicolumn{1}{c|}{\textbf{Models}} & \textbf{F1@M} & {\textbf{F1@5}}  & \textbf{F1@M} & {\textbf{F1@5}} & \textbf{F1@M} & {\textbf{F1@5}} & \textbf{F1@M} & {\textbf{F1@5}} & \textbf{F1@M} & \textbf{F1@5} \\ \hline
\multicolumn{11}{c}{\textbf{Present Keyphrase Generation}}\\ \hline
catSeqTG \cite{chan-etal-2019-neural} & $27.0$ & $22.9$ & $36.6$ & $28.2$ & $29.0$ & $24.6$ & $36.6$ & $29.2$ & --- & ---\\
catSeqTG-2RF1 \cite{chan-etal-2019-neural} & $30.1$ & $25.3$ & $36.9$ & $30.0$ & $32.9$ & $28.7$ & $38.6$ & $32.1$ & --- & --- \\
ExHiRD-h \cite{chen-etal-2020-exclusive} & $29.1_3$ & $25.3_4$ & $34.7_4$ & $28.6_4$ & $33.5_{17}$ & $28.4_{15}$ & $37.4_0$ & $31.1_1$ & --- & ---\\
Transformer \cite{ye-etal-2021-one2set} & $32.5_6$ & $28.1_5$ & $36.5_5$ & $31.5_8$ & $32.5_{15}$ & $28.7_{14}$ & $37.7_1$ & $33.2_1$ & --- & ---\\
SetTrans \cite{ye-etal-2021-one2set} * & $32.4_3$ & $28.5_3$ & $36.4_{12}$ & $32.6_{12}$ & $35.7_{13}$ & $33.1_{20}$ & $39.2_4$ & $35.8_5$ & $54.8$ & ---\\
KPD-A \cite{ChowdhuryPKC22} * & $30.6_3$ & $25.7_{3}$ &  $35.3_6$ & $29.5_7$ & $34.4_5$ & $30.3_7$ & $39.6_2$ & $33.9_3$ & $55.5$ & ---\\
Diversity Heads \cite{thomas-vajjala-2024-improving} & $32.1$ & --- & $37.4$ & --- & $39.6$ & --- & $41.7$ & --- & $\mathbf{56.3}$ & ---\\
UniKeyphrase \cite{wu-etal-2021-unikeyphrase} * & $31.1$ & $29.0$ & --- & --- & $\mathbf{40.9}$ & $\mathbf{41.6}$ & $\mathbf{42.8}$ & $40.8$ & $34.5$ & ---\\
PromptKP \cite{Wu_Ma_Liu_Chen_Nie_2022} & $29.4$ & $26.0$ & --- & --- & $35.6$ & $32.9$ & $35.5$ & $35.1$ & --- & ---\\
SciBART-large \cite{wu-etal-2023-rethinking-model} & $40.2$ & --- & $35.2$ & --- & $34.1$ & --- & $43.1$ & ---& --- & ---\\ 
SimCKP \cite{choi-etal-2023-simckp} & $35.8_8$ & $35.6_6$ & $\mathbf{40.5_8}$ & $\mathbf{40.5_8}$ & $38.6_4$ & $38.7_2$ & $42.7_1$ & $\mathbf{42.6_1}$ & --- & ---\\
ChatGPT TP4 \cite{song2023chatgpt} & $39.3$ & $32.2$ & $16.3$ & $17.0$ & $21.2$ & $23.3$ & $13.6$ & $16.0$ & --- & ---\\
\hline
\multicolumn{11}{l}{\textbf{Ours}}\\ \hline
Llama-3 Multi-sampling & $49.9$ & $45.6$ & $31.8$ & $33.6$ & $38.0$ & $38.1$ & $28.7$ & $32.1$ & $24.7$ & $31.5$\\
Phi-3  Multi-sampling & ${54.7}$ & ${50.9}$ & $25.1$ & $24.7$ & $32.9$ & $30.5$ & $19.7$ & $20.4$ & $12.0$ & $11.9$\\
GPT-4o &  $\mathbf{58.2}$ &  $\mathbf{52.8}$ & $25.9$ & $25.6$ & $32.7$ & $31.4$ & $20.0$ & $21.7 $& $12.1$ & $12.4$ \\
\hline
\multicolumn{11}{c}{\textbf{Absent Keyphrase Generation}}\\ \hline
catSeqTG \cite{chan-etal-2019-neural} & $1.1$ & $0.5$ & $3.4$ & $1.8$ & $2.7$ & $1.9$ & $3.2$ & $1.5$ & --- & ---\\
catSeqTG-2RF1 \cite{chan-etal-2019-neural} & $2.1$ & $1.2$ & $5.3$ & $3.0$ & $3.0$ & $2.1$ & $5.0$ & $2.7$ & --- & ---\\
ExHiRD-h \cite{chen-etal-2020-exclusive} & $2.2_3$ & $1.1_1$ & $4.3_6$ & $2.2_3$ & $2.5_6$ & $1.7_4$ & $3.2_0$ & $1.6_0$ & --- & ---\\
Transformer \cite{ye-etal-2021-one2set} & $1.9_4$ & $1.0_2$ & $6.0_4$ & $3.2_1$ & $2.3_3$ & $2.0_5$ & $4.6_1$ & $2.3_1$ & --- & ---\\
SetTrans \cite{ye-etal-2021-one2set} * & $3.4_3$ & $2.1_1$ & $7.3_{11}$ & $4.7_7$ & $3.4_5$ & $2.6_3$ & $5.8_3$ & $3.6_2$ & $41.2$ & ---\\
KPD-A \cite{ChowdhuryPKC22} * & $3.2_{2}$ & $2.1_{1}$ & $7.2_{7}$ & $4.6_{4}$ & $\mathbf{4.7_{1}}$ & $3.6_{1}$ & $6.6_{1}$ & $4.2_{1}$ & $42.6$ & ---\\
Diversity Heads \cite{thomas-vajjala-2024-improving} & $1.2$ & --- & $7.6$ & --- & $4.2$ & --- & $7.8$ & --- &$\mathbf{44.1}$ & ---\\
UniKeyphrase \cite{wu-etal-2021-unikeyphrase} * & $2.9$ & $2.9$ & --- & --- & $3.2$ & $3.0$ & $4.7$ & $4.7$ & $20.8$ & ---\\
PromptKP \cite{Wu_Ma_Liu_Chen_Nie_2022} & $2.2$ & $1.7$ & --- & --- & $3.2$ & $2.8$ & $4.2$ & $3.2$ & --- & ---\\
SciBART-large \cite{wu-etal-2023-rethinking-model} & $3.6$ & --- & $8.6$ & --- & $4.0$ & --- & $7.6$ & --- & --- & ---\\ 
SimCKP \cite{choi-etal-2023-simckp} & $3.5_3$ & $3.3_2$ & $\mathbf{8.9_0}$ & $\mathbf{7.8_1}$ & $\mathbf{4.7_6}$ & $\mathbf{4.0_2}$ & $\mathbf{8.0_1}$ & $\mathbf{7.3_2}$ & --- & --- \\
ChatGPT TP4 \cite{song2023chatgpt} & $4.1$ & $3.0$ & $1.5$ & $1.1$ & $0.5$ & $0.4$ & $3.9$ & $3.8$ & --- & ---\\
\hline
\multicolumn{11}{l}{\textbf{Ours}}\\ \hline
Llama-3  Multi-sampling & $8.5$ & $7.6$ & $5.9$ & $5.1$ & $3.9$ & $3.6$ & $5.4$ & $4.9$ & $5.3$ & $5.0$\\
Phi-3  Multi-sampling & ${9.3}$ & ${9.0}$ & $2.1$ & $2.1$ & $2.5$ & $2.7$ & $2.0$ & $2.0$ & $0.6$ & $0.7$\\
GPT-4o & $\mathbf{11.7}$ & $\mathbf{10.8}$ & $3.5$ & $3.1$ & $3.7$ & $3.6$ & $2.6$ & $2.6$ & $0.6$ &$ 0.6$ \\
\hline
\end{tabular}
\caption{We compare the performance of our models with various prior works (results from prior works are copied from the corresponding citations; the citations here indicate the source of the results and not necessarily the original work presenting the relevant methods). * Indicates that the kptimes result are taken from \cite{thomas-vajjala-2024-improving} rather than the corresponding citation. Llama3/Phi3/GPT 4o Multisample denotes Llama3/Phi3/GPT 4o multisample (n=10) results with frequency-based ordering and aggregation. KPD-A denotes SetTrans with Greedy Search + KPDrop-A. For brevity, we only present the greedy search results of Diversity Heads \cite{thomas-vajjala-2024-improving} and TP4 prompt style for ChatGPT. SciBART-large indicates the result of (SciBART-large+TAPT+DESEL in \citet{wu-etal-2023-rethinking-model}). $91_{1}$ denotes $91 \pm 0.1$. } 
\label{tb:comparisons}
\vspace{-1mm}
\end{table*}
\subsection{Multi-Sampling Results (RQ3)}
\label{sec:multisampling_results}
In Table \ref{tb:multisampling}, we present the results of multi-sampling with various aggregation strategies. As we would expect, simple union does not help much, and often harms the performance because it removes all order information (which is relevant). Because of our dynamic keyphrase number selection strategy, the order is relevant even for @M metrics. 
Union Concat, Union Interleaf, and Frequency Order are the three best contenders for multi-sampling aggregation.  Among the three, Frequence Order-based aggregation consistently shows the best performance; particularly, on absent keyphrase generation for the open-source models. Overall, we find that the best aggregation methods with multi-sampling significantly improve the performance of LLMs over the baseline. As such, the answer to RQ3 leans towards an affirmation. 
In Appendix Table \ref{tb:absent_recall}, we also show how well absent keyphrases are recalled for the multi-sampling-based approaches compared to the baseline. 

\subsection{Comparison with Prior Works}
As can be seen in Table \ref{tb:comparisons}, our best approaches are competitive against many of the earlier works. Our LLM-based models tend to generate high number of keyphrases which is well suited for Inspec (which also has a high number of keyphrases in the ground truth). As such, our model excels and achieves state of the art in Inspec. In other cases, the overgeneration can become a detriment leading to lower precision when ground truth keyphrases are of fewer numbers. Regardless, our models still remain competitive against many of the prior models in scientific documents. This is especially impressive because this performance is completely zero-shot without any fine-tuning, unlike most prior works. Interestingly, the LLM-based models seem to perform quite poorly in the news domain (KP-Times) compared to others. The gap is particularly high in absent keyphrase generation for KP-Times. Thus, it appears that the LLM-based models, in a zero-shot context, are better biased towards scientific keyphrase extraction, rather than KP-Times-style news domain. 



\section{Additional Analyses}

In Appendix Table \ref{tb:absent_recall} we show the recall of absent keyphrases at higher top-ks. In Appendix \ref{sec:beam_search}, we present result of using multi-sampling aggregations on beam search generations as opposed to independent random sampling. As can be seen independent-sampling based multi-sampling generally outperforms beam search while being more cost-efficient. In Appendix \ref{sec:qualitative_analyses}, we provide qualitative analyses of generated keyphrases. In brief, we find that, under zero-shot prompts, models are biased towards producing high number of longer (multi-word) keyphrases. Inspec best fits this pattern, and thus we find LLMs to ace on Inspec. Whereas KPTimes tend to have short keyphrases and in fewer numbers - potentially a reason for the struggle of zero-shot LLMs in KPTimes.

\section{Related Work}

Identifying keyphrase from a document is a longstanding task and has been well studied in the literature using both supervised, semi-supervised, and unsupervised approaches \cite{PatelC21,PatelCWG20,ParkC20,ChowdhuryCC19,PatelC19,ye-wang-2018-semi,FlorescuC17,HasanN14,GollapalliC14,BougouinBD13,MihalceaT04}. However, with the surge of deep learning models, the attention has shifted towards generative models particularly because of their capability to generate absent keyphrases \cite{wu-etal-2024-leveraging,GargCC23,garg-etal-2022-keyphrase,ChowdhuryPKC22,meng-etal-2017-deep}. 
Many recent works for keyphrase generation have also explored the seq2seq models with no pre-training \cite{meng-etal-2017-deep, chen-etal-2018-keyphrase, ye-wang-2018-semi,chan-etal-2019-neural, swaminathan-etal-2020-preliminary, chen-etal-2020-exclusive, ye-etal-2021-one2set, ye-etal-2021-heterogeneous, Huang_Xu_Jiao_Zu_Zhang_2021,choi-etal-2023-simckp,thomas-vajjala-2024-improving} or pre-trained seq2seq models (e.g., BART) for generating both absent and present keyphrases \cite{DBLP:journals/corr/abs-2004-10462, wu-etal-2021-unikeyphrase, mayank-learning, wu-constrained, wu2022pre, garg-etal-2022-keyphrase,ChowdhuryPKC22,madaan-etal-2022-conditional,wu-etal-2023-rethinking-model,wu-etal-2024-leveraging}. 

More recently, a few works have started to explore decoder-only LLMs for keyphrase generation and extraction \cite{WangLLMs,maragheh2023llmtake,song2023large,song2023chatgpt, martinez2023chatgpt,wu-etal-2024-kpeval}. In our paper, we explore LLMs using novel strategies such as ``specialist prompts'', task-specific instructions, and multi-sampling, and contrast them with many of the above works. 


\vspace{-1mm}
\section{Conclusion and Future Work}
\vspace{-1mm}
In this paper, we addressed three core research questions for keyphrase generation: the effectiveness of specialist prompting for present and absent keyphrases (RQ1), the impact of additional instructions for length and order control (RQ2), and the benefits of multi-sampling for improving keyphrase generation (RQ3). For RQ1, we found that the specialist prompts for present and absent keyphrases did not consistently outperform a simple baseline prompt. 
In terms of RQ2, introducing additional instructions for order and length control yielded mixed results. While length control showed some promise in improving present keyphrase extraction for specific datasets, the overall performance gains were inconsistent across both present and absent keyphrase generation. 
The most promising findings of our paper came from our exploration of RQ3 — the impact of multi-sampling and aggregation. Simple union proved insufficient due to its inability to preserve keyphrase order, which is crucial for certain evaluation metrics like F$_1$@5. However, more sophisticated aggregation techniques, such as Union Concatenation, Union Interleaf, and especially Frequency Order, showed significant improvements in keyphrase generation, particularly for absent keyphrases. Frequency Order, in particular, provided the most consistent results and outperformed the baseline across various settings. 

Our multi-sampling aggregation strategies are also model-agnostic and can work with earlier established KPG models. We leave potential to augment earlier model strategies with multi-sampling aggregation for future work.

\section{Limitations}
This work focuses on zero-shot prompting; however, the effectiveness of few-shot prompting, and parameter-efficient fine-tuning for KPG are also relevant questions that are yet unanswered in this paper. Moreover, alternative evaluation schemes to better judge  LLM's capacities such as KPEval \cite{wu-etal-2024-kpeval} are yet to be explored.
Despite these limitations, we believe our LLM-based methods show promise and offer a strong foundation for future work in LLM-based keyphrase generation.

\section*{Acknowledgements} 
This research is funded in part by NSF CAREER
award 1802358, NSF CRI award 1823292, NSF IIS award 2107518, and an award from Discovery Partners Institute (DPI) at the University of Illinois Chicago.
Any opinions, findings, and conclusions expressed
here are those of the authors and do not necessarily
reflect the views of NSF or DPI. 




\bibliography{acl}

\begin{thebibliography}{65}
\providecommand{\natexlab}[1]{#1}

\bibitem[{Abdin et~al.(2024)Abdin, Jacobs, Awan, Aneja, Awadallah, Awadalla, Bach, Bahree, Bakhtiari, Behl et~al.}]{abdin2024phi}
Marah Abdin, Sam~Ade Jacobs, Ammar~Ahmad Awan, Jyoti Aneja, Ahmed Awadallah, Hany Awadalla, Nguyen Bach, Amit Bahree, Arash Bakhtiari, Harkirat Behl, et~al. 2024.
\newblock Phi-3 technical report: A highly capable language model locally on your phone.
\newblock \emph{arXiv preprint arXiv:2404.14219}.

\bibitem[{Abu-Jbara and Radev(2011)}]{abu-jbara-radev-2011-coherent}
Amjad Abu-Jbara and Dragomir Radev. 2011.
\newblock \href {https://www.aclweb.org/anthology/P11-1051} {Coherent citation-based summarization of scientific papers}.
\newblock In \emph{ACL}, pages 500--509. ACL.

\bibitem[{Achiam et~al.(2023)Achiam, Adler, Agarwal, Ahmad, Akkaya, Aleman, Almeida, Altenschmidt, Altman, Anadkat et~al.}]{achiam2023gpt}
Josh Achiam, Steven Adler, Sandhini Agarwal, Lama Ahmad, Ilge Akkaya, Florencia~Leoni Aleman, Diogo Almeida, Janko Altenschmidt, Sam Altman, Shyamal Anadkat, et~al. 2023.
\newblock Gpt-4 technical report.
\newblock \emph{arXiv preprint arXiv:2303.08774}.

\bibitem[{Al{-}Zaidy et~al.(2019)Al{-}Zaidy, Caragea, and Giles}]{AlzaidyCG19}
Rabah~A. Al{-}Zaidy, Cornelia Caragea, and C.~Lee Giles. 2019.
\newblock \href {https://doi.org/10.1145/3308558.3313642} {Bi-lstm-crf sequence labeling for keyphrase extraction from scholarly documents}.
\newblock In \emph{The World Wide Web Conference, {WWW} 2019, San Francisco, CA, USA, May 13-17, 2019}, pages 2551--2557. {ACM}.

\bibitem[{Augenstein et~al.(2017)Augenstein, Das, Riedel, Vikraman, and McCallum}]{augenstein-etal-2017-semeval}
Isabelle Augenstein, Mrinal Das, Sebastian Riedel, Lakshmi Vikraman, and Andrew McCallum. 2017.
\newblock \href {https://doi.org/10.18653/v1/S17-2091} {{S}em{E}val 2017 task 10: {S}cience{IE} - extracting keyphrases and relations from scientific publications}.
\newblock In \emph{Proceedings of the 11th International Workshop on Semantic Evaluation ({S}em{E}val-2017)}, pages 546--555, Vancouver, Canada. Association for Computational Linguistics.

\bibitem[{Bennani-Smires et~al.(2018)Bennani-Smires, Musat, Hossmann, Baeriswyl, and Jaggi}]{bennani-smires-etal-2018-simple}
Kamil Bennani-Smires, Claudiu Musat, Andreea Hossmann, Michael Baeriswyl, and Martin Jaggi. 2018.
\newblock \href {https://doi.org/10.18653/v1/K18-1022} {Simple unsupervised keyphrase extraction using sentence embeddings}.
\newblock In \emph{Proceedings of the 22nd Conference on Computational Natural Language Learning}, pages 221--229, Brussels, Belgium. Association for Computational Linguistics.

\bibitem[{Boudin et~al.(2020)Boudin, Gallina, and Aizawa}]{boudin2020keyphrase}
Florian Boudin, Ygor Gallina, and Akiko Aizawa. 2020.
\newblock Keyphrase generation for scientific document retrieval.
\newblock In \emph{Proceedings of the 58th Annual Meeting of the Association for Computational Linguistics}, pages 1118--1126.

\bibitem[{Bougouin et~al.(2013)Bougouin, Boudin, and Daille}]{BougouinBD13}
Adrien Bougouin, Florian Boudin, and B{\'{e}}atrice Daille. 2013.
\newblock \href {https://aclanthology.org/I13-1062/} {Topicrank: Graph-based topic ranking for keyphrase extraction}.
\newblock In \emph{Sixth International Joint Conference on Natural Language Processing, {IJCNLP} 2013, Nagoya, Japan, October 14-18, 2013}, pages 543--551. Asian Federation of Natural Language Processing / {ACL}.

\bibitem[{Chan et~al.(2019)Chan, Chen, Wang, and King}]{chan-etal-2019-neural}
Hou~Pong Chan, Wang Chen, Lu~Wang, and Irwin King. 2019.
\newblock \href {https://doi.org/10.18653/v1/P19-1208} {Neural keyphrase generation via reinforcement learning with adaptive rewards}.
\newblock In \emph{Proceedings of the 57th Annual Meeting of the Association for Computational Linguistics}, pages 2163--2174, Florence, Italy. Association for Computational Linguistics.

\bibitem[{Chen et~al.(2018)Chen, Zhang, Wu, Yan, and Li}]{chen-etal-2018-keyphrase}
Jun Chen, Xiaoming Zhang, Yu~Wu, Zhao Yan, and Zhoujun Li. 2018.
\newblock \href {https://doi.org/10.18653/v1/D18-1439} {Keyphrase generation with correlation constraints}.
\newblock In \emph{{EMNLP}}, pages 4057--4066.

\bibitem[{Chen et~al.(2020)Chen, Chan, Li, and King}]{chen-etal-2020-exclusive}
Wang Chen, Hou~Pong Chan, Piji Li, and Irwin King. 2020.
\newblock \href {https://doi.org/10.18653/v1/2020.acl-main.103} {Exclusive hierarchical decoding for deep keyphrase generation}.
\newblock In \emph{Proceedings of the 58th Annual Meeting of the Association for Computational Linguistics}, pages 1095--1105, Online. Association for Computational Linguistics.

\bibitem[{Choi et~al.(2023)Choi, Gwak, Kim, Kim, and Choo}]{choi-etal-2023-simckp}
Minseok Choi, Chaeheon Gwak, Seho Kim, Si~Kim, and Jaegul Choo. 2023.
\newblock \href {https://doi.org/10.18653/v1/2023.findings-emnlp.199} {{S}im{CKP}: Simple contrastive learning of keyphrase representations}.
\newblock In \emph{Findings of the Association for Computational Linguistics: EMNLP 2023}, pages 3003--3015, Singapore. Association for Computational Linguistics.

\bibitem[{Chowdhury et~al.(2019)Chowdhury, Caragea, and Caragea}]{ChowdhuryCC19}
Jishnu~Ray Chowdhury, Cornelia Caragea, and Doina Caragea. 2019.
\newblock \href {https://doi.org/10.1145/3308558.3313696} {Keyphrase extraction from disaster-related tweets}.
\newblock In \emph{The World Wide Web Conference, {WWW} 2019, San Francisco, CA, USA, May 13-17, 2019}, pages 1555--1566. {ACM}.

\bibitem[{Chowdhury et~al.(2022)Chowdhury, Park, Kundu, and Caragea}]{ChowdhuryPKC22}
Jishnu~Ray Chowdhury, Seoyeon Park, Tuhin Kundu, and Cornelia Caragea. 2022.
\newblock \href {https://doi.org/10.18653/V1/2022.FINDINGS-EMNLP.357} {{KPDROP:} improving absent keyphrase generation}.
\newblock In \emph{Findings of the Association for Computational Linguistics: {EMNLP} 2022, Abu Dhabi, United Arab Emirates, December 7-11, 2022}, pages 4853--4870. Association for Computational Linguistics.

\bibitem[{Dubey et~al.(2024)Dubey, Jauhri, Pandey, Kadian, Al-Dahle, Letman, Mathur, Schelten, Yang, Fan et~al.}]{dubey2024llama}
Abhimanyu Dubey, Abhinav Jauhri, Abhinav Pandey, Abhishek Kadian, Ahmad Al-Dahle, Aiesha Letman, Akhil Mathur, Alan Schelten, Amy Yang, Angela Fan, et~al. 2024.
\newblock The llama 3 herd of models.
\newblock \emph{arXiv preprint arXiv:2407.21783}.

\bibitem[{Florescu and Caragea(2017)}]{FlorescuC17}
Corina Florescu and Cornelia Caragea. 2017.
\newblock \href {https://doi.org/10.18653/V1/P17-1102} {Positionrank: An unsupervised approach to keyphrase extraction from scholarly documents}.
\newblock In \emph{Proceedings of the 55th Annual Meeting of the Association for Computational Linguistics, {ACL} 2017, Vancouver, Canada, July 30 - August 4, Volume 1: Long Papers}, pages 1105--1115. Association for Computational Linguistics.

\bibitem[{Gallina et~al.(2019)Gallina, Boudin, and Daille}]{gallina2019kptimes}
Ygor Gallina, Florian Boudin, and Beatrice Daille. 2019.
\newblock Kptimes: A large-scale dataset for keyphrase generation on news documents.
\newblock \emph{arXiv preprint arXiv:1911.12559}.

\bibitem[{Garg et~al.(2023)Garg, Chowdhury, and Caragea}]{GargCC23}
Krishna Garg, Jishnu~Ray Chowdhury, and Cornelia Caragea. 2023.
\newblock \href {https://doi.org/10.18653/V1/2023.FINDINGS-ACL.534} {Data augmentation for low-resource keyphrase generation}.
\newblock In \emph{Findings of the Association for Computational Linguistics: {ACL} 2023, Toronto, Canada, July 9-14, 2023}, pages 8442--8455. Association for Computational Linguistics.

\bibitem[{Garg et~al.(2022)Garg, Ray~Chowdhury, and Caragea}]{garg-etal-2022-keyphrase}
Krishna Garg, Jishnu Ray~Chowdhury, and Cornelia Caragea. 2022.
\newblock \href {https://aclanthology.org/2022.findings-emnlp.427} {Keyphrase generation beyond the boundaries of title and abstract}.
\newblock In \emph{Findings of the Association for Computational Linguistics: EMNLP 2022}, pages 5809--5821, Abu Dhabi, United Arab Emirates. Association for Computational Linguistics.

\bibitem[{Gollapalli and Caragea(2014)}]{GollapalliC14}
Sujatha~Das Gollapalli and Cornelia Caragea. 2014.
\newblock \href {https://doi.org/10.1609/AAAI.V28I1.8946} {Extracting keyphrases from research papers using citation networks}.
\newblock In \emph{Proceedings of the Twenty-Eighth {AAAI} Conference on Artificial Intelligence, July 27 -31, 2014, Qu{\'{e}}bec City, Qu{\'{e}}bec, Canada}, pages 1629--1635. {AAAI} Press.

\bibitem[{Hasan and Ng(2014)}]{HasanN14}
Kazi~Saidul Hasan and Vincent Ng. 2014.
\newblock \href {https://doi.org/10.3115/V1/P14-1119} {Automatic keyphrase extraction: {A} survey of the state of the art}.
\newblock In \emph{Proceedings of the 52nd Annual Meeting of the Association for Computational Linguistics, {ACL} 2014, June 22-27, 2014, Baltimore, MD, USA, Volume 1: Long Papers}, pages 1262--1273. The Association for Computer Linguistics.

\bibitem[{Huang et~al.(2021)Huang, Xu, Jiao, Zu, and Zhang}]{Huang_Xu_Jiao_Zu_Zhang_2021}
Xiaoli Huang, Tongge Xu, Lvan Jiao, Yueran Zu, and Youmin Zhang. 2021.
\newblock \href {https://ojs.aaai.org/index.php/AAAI/article/view/17546} {Adaptive beam search decoding for discrete keyphrase generation}.
\newblock \emph{Proceedings of the AAAI Conference on Artificial Intelligence}, 35(14):13082--13089.

\bibitem[{Iyer et~al.(2022)Iyer, Lin, Pasunuru, Mihaylov, Simig, Yu, Shuster, Wang, Liu, Koura et~al.}]{iyer2022opt}
Srinivasan Iyer, Xi~Victoria Lin, Ramakanth Pasunuru, Todor Mihaylov, Daniel Simig, Ping Yu, Kurt Shuster, Tianlu Wang, Qing Liu, Punit~Singh Koura, et~al. 2022.
\newblock Opt-iml: Scaling language model instruction meta learning through the lens of generalization.
\newblock \emph{arXiv preprint arXiv:2212.12017}.

\bibitem[{Jones and Staveley(1999)}]{jones1999phrasier}
Steve Jones and Mark~S Staveley. 1999.
\newblock Phrasier: a system for interactive document retrieval using keyphrases.
\newblock In \emph{Proceedings of the 22nd annual international ACM SIGIR conference on Research and development in information retrieval}, pages 160--167.

\bibitem[{Joshi et~al.(2023)Joshi, Balachandran, Saldanha, Glenski, Volkova, and Tsvetkov}]{joshi2023unsupervisedkeyphraseextractioninterpretable}
Rishabh Joshi, Vidhisha Balachandran, Emily Saldanha, Maria Glenski, Svitlana Volkova, and Yulia Tsvetkov. 2023.
\newblock \href {https://arxiv.org/abs/2203.07640} {Unsupervised keyphrase extraction via interpretable neural networks}.
\newblock \emph{Preprint}, arXiv:2203.07640.

\bibitem[{Kim et~al.(2010)Kim, Medelyan, Kan, and Baldwin}]{kim-etal-2010-semeval}
Su~Nam Kim, Olena Medelyan, Min-Yen Kan, and Timothy Baldwin. 2010.
\newblock \href {https://aclanthology.org/S10-1004} {{S}em{E}val-2010 task 5 : Automatic keyphrase extraction from scientific articles}.
\newblock In \emph{Proceedings of the 5th International Workshop on Semantic Evaluation}, pages 21--26, Uppsala, Sweden. Association for Computational Linguistics.

\bibitem[{Krapivin et~al.(2009)Krapivin, Autaeu, and Marchese}]{Krapivin2009LargeDF}
Mikalai Krapivin, Aliaksandr Autaeu, and Maurizio Marchese. 2009.
\newblock \href {https://api.semanticscholar.org/CorpusID:60540897} {Large dataset for keyphrases extraction}.

\bibitem[{Kulkarni et~al.(2022)Kulkarni, Mahata, Arora, and Bhowmik}]{mayank-learning}
Mayank Kulkarni, Debanjan Mahata, Ravneet Arora, and Rajarshi Bhowmik. 2022.
\newblock \href {https://doi.org/10.18653/v1/2022.findings-naacl.67} {Learning rich representation of keyphrases from text}.
\newblock In \emph{Findings of the Association for Computational Linguistics: NAACL 2022}, pages 891--906, Seattle, United States. Association for Computational Linguistics.

\bibitem[{Liu et~al.(2020)Liu, Lin, and Wang}]{DBLP:journals/corr/abs-2004-10462}
Rui Liu, Zheng Lin, and Weiping Wang. 2020.
\newblock \href {https://arxiv.org/abs/2004.10462} {Keyphrase prediction with pre-trained language model}.
\newblock \emph{CoRR}, abs/2004.10462.

\bibitem[{Madaan et~al.(2022)Madaan, Rajagopal, Tandon, Yang, and Bosselut}]{madaan-etal-2022-conditional}
Aman Madaan, Dheeraj Rajagopal, Niket Tandon, Yiming Yang, and Antoine Bosselut. 2022.
\newblock \href {https://aclanthology.org/2022.emnlp-main.324} {Conditional set generation using seq2seq models}.
\newblock In \emph{Proceedings of the 2022 Conference on Empirical Methods in Natural Language Processing}, pages 4874--4896, Abu Dhabi, United Arab Emirates. Association for Computational Linguistics.

\bibitem[{Maragheh et~al.(2023)Maragheh, Fang, Irugu, Parikh, Cho, Xu, Sukumar, Patel, Korpeoglu, Kumar, and Achan}]{maragheh2023llmtake}
Reza~Yousefi Maragheh, Chenhao Fang, Charan~Chand Irugu, Parth Parikh, Jason Cho, Jianpeng Xu, Saranyan Sukumar, Malay Patel, Evren Korpeoglu, Sushant Kumar, and Kannan Achan. 2023.
\newblock \href {https://arxiv.org/abs/2312.00909} {Llm-take: Theme aware keyword extraction using large language models}.
\newblock \emph{Preprint}, arXiv:2312.00909.

\bibitem[{Mart{\'\i}nez-Cruz et~al.(2023)Mart{\'\i}nez-Cruz, L{\'o}pez-L{\'o}pez, and Portela}]{martinez2023chatgpt}
Roberto Mart{\'\i}nez-Cruz, Alvaro~J L{\'o}pez-L{\'o}pez, and Jos{\'e} Portela. 2023.
\newblock Chatgpt vs state-of-the-art models: a benchmarking study in keyphrase generation task.
\newblock \emph{arXiv preprint arXiv:2304.14177}.

\bibitem[{Meng et~al.(2017)Meng, Zhao, Han, He, Brusilovsky, and Chi}]{meng-etal-2017-deep}
Rui Meng, Sanqiang Zhao, Shuguang Han, Daqing He, Peter Brusilovsky, and Yu~Chi. 2017.
\newblock \href {https://doi.org/10.18653/v1/P17-1054} {Deep keyphrase generation}.
\newblock In \emph{Proceedings of the 55th Annual Meeting of the Association for Computational Linguistics (Volume 1: Long Papers)}, pages 582--592, Vancouver, Canada. Association for Computational Linguistics.

\bibitem[{Mihalcea and Tarau(2004)}]{MihalceaT04}
Rada Mihalcea and Paul Tarau. 2004.
\newblock \href {https://aclanthology.org/W04-3252/} {Textrank: Bringing order into text}.
\newblock In \emph{Proceedings of the 2004 Conference on Empirical Methods in Natural Language Processing , {EMNLP} 2004, {A} meeting of SIGDAT, a Special Interest Group of the ACL, held in conjunction with {ACL} 2004, 25-26 July 2004, Barcelona, Spain}, pages 404--411. {ACL}.

\bibitem[{Moulin and Mihcak(2002)}]{moulin2002framework}
Pierre Moulin and Mehmet~Kivan{\c{c}} Mihcak. 2002.
\newblock A framework for evaluating the data-hiding capacity of image sources.
\newblock \emph{IEEE Transactions on Image Processing}, 11(9):1029--1042.

\bibitem[{News(2019)}]{nbcnews2019}
NBC News. 2019.
\newblock \href {https://www.nbcnews.com/news/us-news/manafort-family-business-defends-name-infamous-cousin-sits-jail-n978646} {Manafort family business defends its name as infamous cousin sits in jail}.
\newblock Accessed: 2025-01-26.

\bibitem[{Park and Caragea(2023)}]{ParkC23}
Seo Park and Cornelia Caragea. 2023.
\newblock \href {https://doi.org/10.18653/V1/2023.EMNLP-MAIN.805} {Multi-task knowledge distillation with embedding constraints for scholarly keyphrase boundary classification}.
\newblock In \emph{Proceedings of the 2023 Conference on Empirical Methods in Natural Language Processing, {EMNLP} 2023, Singapore, December 6-10, 2023}, pages 13026--13042. Association for Computational Linguistics.

\bibitem[{Park and Caragea(2020)}]{ParkC20}
Seoyeon Park and Cornelia Caragea. 2020.
\newblock \href {https://doi.org/10.18653/V1/2020.COLING-MAIN.472} {Scientific keyphrase identification and classification by pre-trained language models intermediate task transfer learning}.
\newblock In \emph{Proceedings of the 28th International Conference on Computational Linguistics, {COLING} 2020, Barcelona, Spain (Online), December 8-13, 2020}, pages 5409--5419. International Committee on Computational Linguistics.

\bibitem[{Patel and Caragea(2019)}]{PatelC19}
Krutarth Patel and Cornelia Caragea. 2019.
\newblock \href {https://doi.org/10.1145/3360901.3364447} {Exploring word embeddings in crf-based keyphrase extraction from research papers}.
\newblock In \emph{Proceedings of the 10th International Conference on Knowledge Capture, {K-CAP} 2019, Marina Del Rey, CA, USA, November 19-21, 2019}, pages 37--44. {ACM}.

\bibitem[{Patel and Caragea(2021)}]{PatelC21}
Krutarth Patel and Cornelia Caragea. 2021.
\newblock \href {https://doi.org/10.18653/V1/2021.EACL-MAIN.136} {Exploiting position and contextual word embeddings for keyphrase extraction from scientific papers}.
\newblock In \emph{Proceedings of the 16th Conference of the European Chapter of the Association for Computational Linguistics: Main Volume, {EACL} 2021, Online, April 19 - 23, 2021}, pages 1585--1591. Association for Computational Linguistics.

\bibitem[{Patel et~al.(2020)Patel, Caragea, Wu, and Giles}]{PatelCWG20}
Krutarth Patel, Cornelia Caragea, Jian Wu, and C.~Lee Giles. 2020.
\newblock \href {https://doi.org/10.1007/978-3-030-59618-7\_12} {Keyphrase extraction in scholarly digital library search engines}.
\newblock In \emph{Web Services - {ICWS} 2020 - 27th International Conference, Held as Part of the Services Conference Federation, {SCF} 2020, Honolulu, HI, USA, September 18-20, 2020, Proceedings}, volume 12406 of \emph{Lecture Notes in Computer Science}, pages 179--196. Springer.

\bibitem[{Reips and Neuhaus(2002)}]{reips2002wextor}
Ulf-Dietrich Reips and Christoph Neuhaus. 2002.
\newblock Wextor: A web-based tool for generating and visualizing experimental designs and procedures.
\newblock \emph{Behavior Research Methods, Instruments, \& Computers}, 34:234--240.

\bibitem[{Song et~al.(2006)Song, Song, Allen, and Obradovic}]{10.1145/1141753.1141800}
Min Song, Il~Yeol Song, Robert~B. Allen, and Zoran Obradovic. 2006.
\newblock \href {https://doi.org/10.1145/1141753.1141800} {Keyphrase extraction-based query expansion in digital libraries}.
\newblock In \emph{Proceedings of the 6th ACM/IEEE-CS Joint Conference on Digital Libraries}, JCDL '06, page 202–209, New York, NY, USA. Association for Computing Machinery.

\bibitem[{Song et~al.(2023{\natexlab{a}})Song, Geng, Yao, Lu, Feng, and Jing}]{song2023large}
Mingyang Song, Xuelian Geng, Songfang Yao, Shilong Lu, Yi~Feng, and Liping Jing. 2023{\natexlab{a}}.
\newblock Large language models as zero-shot keyphrase extractor: A preliminary empirical study.
\newblock \emph{arXiv preprint arXiv:2312.15156}.

\bibitem[{Song et~al.(2023{\natexlab{b}})Song, Jiang, Shi, Yao, Lu, Feng, Liu, and Jing}]{song2023chatgpt}
Mingyang Song, Haiyun Jiang, Shuming Shi, Songfang Yao, Shilong Lu, Yi~Feng, Huafeng Liu, and Liping Jing. 2023{\natexlab{b}}.
\newblock Is chatgpt a good keyphrase generator? a preliminary study.
\newblock \emph{arXiv preprint arXiv:2303.13001}.

\bibitem[{Sterckx et~al.(2016)Sterckx, Caragea, Demeester, and Develder}]{SterckxCDD16}
Lucas Sterckx, Cornelia Caragea, Thomas Demeester, and Chris Develder. 2016.
\newblock \href {https://doi.org/10.18653/V1/D16-1198} {Supervised keyphrase extraction as positive unlabeled learning}.
\newblock In \emph{Proceedings of the 2016 Conference on Empirical Methods in Natural Language Processing, {EMNLP} 2016, Austin, Texas, USA, November 1-4, 2016}, pages 1924--1929. The Association for Computational Linguistics.

\bibitem[{Swaminathan et~al.(2020)Swaminathan, Zhang, Mahata, Gosangi, Shah, and Stent}]{swaminathan-etal-2020-preliminary}
Avinash Swaminathan, Haimin Zhang, Debanjan Mahata, Rakesh Gosangi, Rajiv~Ratn Shah, and Amanda Stent. 2020.
\newblock \href {https://doi.org/10.18653/v1/2020.emnlp-main.645} {A preliminary exploration of {GAN}s for keyphrase generation}.
\newblock In \emph{Proceedings of the 2020 Conference on Empirical Methods in Natural Language Processing (EMNLP)}, pages 8021--8030, Online. Association for Computational Linguistics.

\bibitem[{Thomas and Vajjala(2024)}]{thomas-vajjala-2024-improving}
Edwin Thomas and Sowmya Vajjala. 2024.
\newblock \href {https://doi.org/10.18653/v1/2024.findings-naacl.102} {Improving absent keyphrase generation with diversity heads}.
\newblock In \emph{Findings of the Association for Computational Linguistics: NAACL 2024}, pages 1568--1584, Mexico City, Mexico. Association for Computational Linguistics.

\bibitem[{Times(2014)}]{japantimes2014}
The~Japan Times. 2014.
\newblock \href {https://www.japantimes.co.jp/news/2014/11/24/national/chinese-tourists-step-abe-japanese-tighten-belts/} {Chinese tourists step up as abe, japanese tighten belts}.
\newblock Accessed: 2025-01-26.

\bibitem[{Touvron et~al.(2023)Touvron, Lavril, Izacard, Martinet, Lachaux, Lacroix, Rozière, Goyal, Hambro, Azhar, Rodriguez, Joulin, Grave, and Lample}]{touvron2023llamaopenefficientfoundation}
Hugo Touvron, Thibaut Lavril, Gautier Izacard, Xavier Martinet, Marie-Anne Lachaux, Timothée Lacroix, Baptiste Rozière, Naman Goyal, Eric Hambro, Faisal Azhar, Aurelien Rodriguez, Armand Joulin, Edouard Grave, and Guillaume Lample. 2023.
\newblock \href {https://arxiv.org/abs/2302.13971} {Llama: Open and efficient foundation language models}.
\newblock \emph{Preprint}, arXiv:2302.13971.

\bibitem[{Wang and Cardie(2013)}]{wang2013domain}
Lu~Wang and Claire Cardie. 2013.
\newblock Domain-independent abstract generation for focused meeting summarization.
\newblock In \emph{Proceedings of the 51st Annual Meeting of the Association for Computational Linguistics (Volume 1: Long Papers)}, pages 1395--1405.

\bibitem[{Wang et~al.(2023)Wang, Wei, Schuurmans, Le, Chi, Narang, Chowdhery, and Zhou}]{wang2023selfconsistency}
Xuezhi Wang, Jason Wei, Dale Schuurmans, Quoc~V Le, Ed~H. Chi, Sharan Narang, Aakanksha Chowdhery, and Denny Zhou. 2023.
\newblock \href {https://openreview.net/forum?id=1PL1NIMMrw} {Self-consistency improves chain of thought reasoning in language models}.
\newblock In \emph{The Eleventh International Conference on Learning Representations}.

\bibitem[{Wang et~al.(2024)Wang, Sha, Lin, Feng, Zhu, Wang, Jiao, Huang, Ye, He, Guo, Li, and Liu}]{WangLLMs}
Yang Wang, Zheyi Sha, Kunhai Lin, Chaobing Feng, Kunhong Zhu, Lipeng Wang, Xuewu Jiao, Fei Huang, Chao Ye, Dengwu He, Zhi Guo, Shuanglong Li, and Lin Liu. 2024.
\newblock \href {https://doi.org/10.1145/3589335.3651943} {One-step reach: Llm-based keyword generation for sponsored search advertising}.
\newblock In \emph{Companion Proceedings of the ACM on Web Conference 2024}, WWW '24, page 1604–1608, New York, NY, USA. Association for Computing Machinery.

\bibitem[{Wu et~al.(2023)Wu, Ahmad, and Chang}]{wu-etal-2023-rethinking-model}
Di~Wu, Wasi Ahmad, and Kai-Wei Chang. 2023.
\newblock \href {https://doi.org/10.18653/v1/2023.emnlp-main.410} {Rethinking model selection and decoding for keyphrase generation with pre-trained sequence-to-sequence models}.
\newblock In \emph{Proceedings of the 2023 Conference on Empirical Methods in Natural Language Processing}, pages 6642--6658, Singapore. Association for Computational Linguistics.

\bibitem[{Wu et~al.(2024{\natexlab{a}})Wu, Ahmad, and Chang}]{wu-etal-2024-leveraging}
Di~Wu, Wasi Ahmad, and Kai-Wei Chang. 2024{\natexlab{a}}.
\newblock \href {https://aclanthology.org/2024.lrec-main.1083} {On leveraging encoder-only pre-trained language models for effective keyphrase generation}.
\newblock In \emph{Proceedings of the 2024 Joint International Conference on Computational Linguistics, Language Resources and Evaluation (LREC-COLING 2024)}, pages 12370--12384, Torino, Italia. ELRA and ICCL.

\bibitem[{Wu et~al.(2022{\natexlab{a}})Wu, Ahmad, and Chang}]{wu2022pre}
Di~Wu, Wasi~Uddin Ahmad, and Kai-Wei Chang. 2022{\natexlab{a}}.
\newblock Pre-trained language models for keyphrase generation: A thorough empirical study.
\newblock \emph{arXiv preprint arXiv:2212.10233}.

\bibitem[{Wu et~al.(2022{\natexlab{b}})Wu, Ahmad, Dev, and Chang}]{wu-constrained}
Di~Wu, Wasi~Uddin Ahmad, Sunipa Dev, and Kai{-}Wei Chang. 2022{\natexlab{b}}.
\newblock \href {https://doi.org/10.48550/arXiv.2203.08118} {Representation learning for resource-constrained keyphrase generation}.
\newblock \emph{ArXiv}, abs/2203.08118.

\bibitem[{Wu et~al.(2024{\natexlab{b}})Wu, Yin, and Chang}]{wu-etal-2024-kpeval}
Di~Wu, Da~Yin, and Kai-Wei Chang. 2024{\natexlab{b}}.
\newblock \href {https://doi.org/10.18653/v1/2024.findings-acl.117} {{KPE}val: Towards fine-grained semantic-based keyphrase evaluation}.
\newblock In \emph{Findings of the Association for Computational Linguistics: ACL 2024}, pages 1959--1981, Bangkok, Thailand. Association for Computational Linguistics.

\bibitem[{Wu et~al.(2021)Wu, Liu, Li, Nie, Chen, Zhang, and Wang}]{wu-etal-2021-unikeyphrase}
Huanqin Wu, Wei Liu, Lei Li, Dan Nie, Tao Chen, Feng Zhang, and Di~Wang. 2021.
\newblock \href {https://doi.org/10.18653/v1/2021.findings-acl.73} {{U}ni{K}eyphrase: A unified extraction and generation framework for keyphrase prediction}.
\newblock In \emph{Findings of the Association for Computational Linguistics: ACL-IJCNLP 2021}, pages 825--835, Online. Association for Computational Linguistics.

\bibitem[{Wu et~al.(2022{\natexlab{c}})Wu, Ma, Liu, Chen, and Nie}]{Wu_Ma_Liu_Chen_Nie_2022}
Huanqin Wu, Baijiaxin Ma, Wei Liu, Tao Chen, and Dan Nie. 2022{\natexlab{c}}.
\newblock \href {https://doi.org/10.1609/aaai.v36i10.21402} {Fast and constrained absent keyphrase generation by prompt-based learning}.
\newblock \emph{Proceedings of the AAAI Conference on Artificial Intelligence}, 36(10):11495--11503.

\bibitem[{Ye and Wang(2018)}]{ye-wang-2018-semi}
Hai Ye and Lu~Wang. 2018.
\newblock \href {https://doi.org/10.18653/v1/D18-1447} {Semi-supervised learning for neural keyphrase generation}.
\newblock In \emph{Proceedings of the 2018 Conference on Empirical Methods in Natural Language Processing}, pages 4142--4153, Brussels, Belgium. Association for Computational Linguistics.

\bibitem[{Ye et~al.(2021{\natexlab{a}})Ye, Cai, Gui, and Zhang}]{ye-etal-2021-heterogeneous}
Jiacheng Ye, Ruijian Cai, Tao Gui, and Qi~Zhang. 2021{\natexlab{a}}.
\newblock \href {https://doi.org/10.18653/v1/2021.emnlp-main.213} {Heterogeneous graph neural networks for keyphrase generation}.
\newblock In \emph{Proceedings of the 2021 Conference on Empirical Methods in Natural Language Processing}, pages 2705--2715, Online and Punta Cana, Dominican Republic. Association for Computational Linguistics.

\bibitem[{Ye et~al.(2021{\natexlab{b}})Ye, Gui, Luo, Xu, and Zhang}]{ye-etal-2021-one2set}
Jiacheng Ye, Tao Gui, Yichao Luo, Yige Xu, and Qi~Zhang. 2021{\natexlab{b}}.
\newblock \href {https://doi.org/10.18653/v1/2021.acl-long.354} {{O}ne2{S}et: {G}enerating diverse keyphrases as a set}.
\newblock In \emph{Proceedings of the 59th Annual Meeting of the Association for Computational Linguistics and the 11th International Joint Conference on Natural Language Processing (Volume 1: Long Papers)}, pages 4598--4608, Online. Association for Computational Linguistics.

\bibitem[{Yu and Ng(2018)}]{YuN18}
Yang Yu and Vincent Ng. 2018.
\newblock \href {http://www.lrec-conf.org/proceedings/lrec2018/summaries/871.html} {Improving unsupervised keyphrase extraction using background knowledge}.
\newblock In \emph{Proceedings of the Eleventh International Conference on Language Resources and Evaluation, {LREC} 2018, Miyazaki, Japan, May 7-12, 2018}. European Language Resources Association {(ELRA)}.

\bibitem[{Yuan et~al.(2020)Yuan, Wang, Meng, Thaker, Brusilovsky, He, and Trischler}]{yuan2020one}
Xingdi Yuan, Tong Wang, Rui Meng, Khushboo Thaker, Peter Brusilovsky, Daqing He, and Adam Trischler. 2020.
\newblock One size does not fit all: Generating and evaluating variable number of keyphrases.
\newblock In \emph{Proceedings of the 58th Annual Meeting of the Association for Computational Linguistics}, pages 7961--7975.

\end{thebibliography}

\appendix
\clearpage
\section{Evaluation}
\label{sec:evaluation}
We consider the following standard evaluations for KPG:
\begin{enumerate}
    \item F$_1$@M: This evaluation metric calculates the F$_1$ between all the predicted keyphrases by the model and the ground truth keyphrases. In the case of multi-sampling models, the term ``all the predicted keyphrases" stands for all the keyphrases that remain after selection of the top-$M_{pre}$ present keyphrases and top-$M_{abs}$ absent keyphrases based on the dynamic keyphrase number selection that we discussed before.  
    \item F$_1$@5: This evaluation metric calculates the F$_1$ between the top-$5$ predicted keyphrases by the model and the ground truth keyphrases. Similar to \cite{chan-etal-2019-neural} and others, in case there are less than $5$ predicted keyphrases, we add dummy ones until there are 5 keyphrases.
    \item R@10: This evaluation metric calculates the recall between the top-$10$ predicted keyphrases by the model and the ground truth keyphrases. 
    \item R@Inf: This evaluation metric calculates the recall between all the predicted keyphrases (with no truncation) by the model and the ground truth keyphrases. The difference between @Inf and @M is that in the context of multi-sampling models, for @Inf, we do not truncate the keyphrases based on dynamically determined $M_{pre}$ and $M_{abs}$ values. Otherwise, for any other case, @M and @Inf are equivalent. R@Inf shows the upperbound performance that we can get if we have a perfect selector to select from the raw list of predictions from all samples of a model for any specific input.
\end{enumerate}
For all cases, we calculate the macro-average as is the standard. Following convention, we distinguish between absent and present keyphrases based on whether the lower-cased stemmed (using PorterStemmer) version of keyphrases match with the lowercased stemmed version of the input text. 


\begin{table*}[t]
\centering
\footnotesize
\def\arraystretch{1.2}
\begin{tabular}{l|cc|cc|cc|cc|cc}
\hline
 & \multicolumn{2}{c}{\textbf{Inspec}} & \multicolumn{2}{|c}{\textbf{Krapivin}} & \multicolumn{2}{|c}{\textbf{SemEval}} & \multicolumn{2}{|c}{\textbf{KP20K}} & \multicolumn{2}{|c}{\textbf{KPTimes}} \\  
\multicolumn{1}{c|}{\textbf{Models}} & \textbf{R@10} & \textbf{R@Inf}  & \textbf{R@10} & \textbf{R@Inf} & \textbf{R@10} & \textbf{R@Inf} & \textbf{R@10} & \textbf{R@Inf} & \textbf{R@10} & \textbf{R@Inf} \\ \hline
\multicolumn{11}{l}{\textbf{Llama-3.0 8B Instruct}}\\ \hline
Baseline & 7.6 & 7.6 & 5.5 & 5.5 & 2.5 & 2.5 & 4.2 & 4.2 & 4.9 & 4.9\\
\hdashline
\multicolumn{11}{l}{Multi-sampling (n=10)} \\
\hdashline
Union & 12.2 & \textbf{17.1} & 7.6 & \textbf{11.8} & 3.1 & \textbf{5.0} & 8.8 & \textbf{11.6} & 7.0 & \textbf{14.0}\\
Union Concat & 15.6 & \textbf{17.1} & 9.6 & \textbf{11.8} & 3.5 & \textbf{5.0} & 9.8 & \textbf{11.6} & 10.0 & \textbf{14.0}\\
Union Interleaf & 14.5 & \textbf{17.1} & \textbf{9.8} & \textbf{11.8} & \textbf{3.6} & \textbf{5.0} & 10.1 & \textbf{11.6} & 10.8 & \textbf{14.0}\\
Frequency Order & \textbf{15.2} & \textbf{17.1} & 9.6 & \textbf{11.8} & 3.5 & \textbf{5.0} & \textbf{10.2} & \textbf{11.6} & \textbf{10.9} & \textbf{14.0}\\
\hline
\multicolumn{11}{l}{\textbf{Phi-3.0 3.8B Mini 128K Instruct}}\\ \hline
Baseline & 9.4 & 9.4 & 1.7 & 1.7 & 1.9 & 1.9 & 1.8 & 1.9 & 0.7 & 0.8\\
\hdashline
\multicolumn{11}{l}{Multi-sampling (n=10)} \\
\hdashline
Union &  7.5 & \textbf{23.8} & 1.8 & \textbf{6.9} & 1.7 & \textbf{4.6} & 2.5 & \textbf{7.5} & 0.7 & \textbf{5.2}\\
Union Concat & 16.2 & \textbf{23.8} & 3.4 & \textbf{6.9} & 2.8 & \textbf{4.6} & 3.8 & \textbf{7.5} & 1.2 & \textbf{5.2}\\
Union Interleaf & 15.4 & \textbf{23.8} & 3.4 & \textbf{6.9} & 1.6 & \textbf{4.6} & 4.2 & \textbf{7.5} & 1.3 & \textbf{5.2}\\
Frequency Order & \textbf{19.7} & \textbf{23.8} & \textbf{3.8} & \textbf{6.9} & \textbf{3.0} & \textbf{4.6} & \textbf{4.6} & \textbf{7.5} & \textbf{1.8} & \textbf{5.2}\\
\hline
\end{tabular}
\caption{Recall performance of our multi-sample models for absent keyphrase generation. R indicates recall. @Inf indicates that all keyphrases from all samples for an input is considered without any dynamic @M selection.} 
\label{tb:absent_recall}
\end{table*}

\section{Beam Search}
\label{sec:beam_search}

Beam search is a search algorithm used in sequence generation tasks, aiming to balance between exploration and exploitation. It maintains a set of the \( k \) most probable hypotheses at each step, where \( k \) is the beam width. The model computes a probability distribution over the next token, and at each step, the \( k \) most probable sequences are kept and expanded. Mathematically, given the sequence \( \mathbf{X}_{t-1} = (x_1, x_2, \dots, x_{t-1}) \), the probability of the next token is computed as:
\[
P(x_t | \mathbf{X}_{t-1})
\]
Beam search proceeds by maintaining and expanding the top \( k \) sequences, based on their cumulative probability:
\[
P(\mathbf{x}_t) = \prod_{i=1}^{t} P(x_i | \mathbf{X}_{i-1})
\]
After expanding all sequences, the top \( k \) sequences are retained, and this process repeats until a stopping criterion (e.g., reaching the end token) is met. Following the multi-sampling experiments detailed in Section \ref{sec:multisampling_results}, we applied the same aggregation strategies to evaluate the performance of the beam search strategy. Table \ref{tb:beamsearch} summarizes the results of beam search conducted on the open-source models Llama-3.0 and Phi-3.0, using a beam width of 10. Consistent with our previous experiments, the generation length was constrained to 500 tokens, with all other parameters held constant. The results indicate that the multi-sampling strategy with various aggregation techniques, consistently outperforms the standard beam search approach across all datasets. 

\begin{table}[t]
\centering
\def\arraystretch{1.6}
\resizebox{0.48\textwidth}{!}{%
\begin{tabular}{|c|c|c|ccc|}
\hline
\multirow{2}{*}{\textbf{Dataset}} &
  \multirow{2}{*}{\textbf{Statistics Names}} &
  \multirow{2}{*}{\textbf{\begin{tabular}[c]{@{}c@{}}Original Data\\ (Ground Truth)\end{tabular}}} &
  \multicolumn{3}{c|}{\multirow{2}{*}{\textbf{Statistics of Model Generations}}} \\
                                   &                                             &        & \multicolumn{3}{c|}{}                                         \\ \hline
\multicolumn{1}{|l|}{} &
  \multicolumn{1}{l|}{} &
  \multicolumn{1}{l|}{} &
  \multicolumn{1}{c|}{\textbf{\hspace{0.1cm} Llama-3 \hspace{0.1cm}}} &
  \multicolumn{1}{c|}{\textbf{\hspace{0.1cm}Gpt-4o\hspace{0.1cm}}} &

  \textbf{Phi-3} \\ \hline
\multirow{5}{*}{\textbf{Inspec}} &
  Average words in Title + Abstract &
  121.82 &
  \multicolumn{1}{c|}{\textbf{}} &
  \multicolumn{1}{c|}{\textbf{}} &
  \textbf{} \\
                                   & Average words per present keyphrase         & 2.27   & \multicolumn{1}{c|}{2.00} & \multicolumn{1}{c|}{2.36}  & 2.43 \\
                                   & Average words per absent keyphrase          & 2.52   & \multicolumn{1}{c|}{2.14} & \multicolumn{1}{c|}{2.69}  & 3.07 \\
                                   & Average no. of present keyphrases per input & 7.70   & \multicolumn{1}{c|}{7.91} & \multicolumn{1}{c|}{8.33}  & 7.33 \\
                                   & Average no. of absent keyphrases per input  & 2.15   & \multicolumn{1}{c|}{2.43} & \multicolumn{1}{c|}{3.7}   & 5.26 \\ \hline
\multirow{5}{*}{\textbf{Krapivin}} & Average Words in Title + Abstract           & 180.65 & \multicolumn{1}{c|}{}     & \multicolumn{1}{c|}{}      &      \\
                                   & Average words per present keyphrase         & 2.15   & \multicolumn{1}{c|}{2.10} & \multicolumn{1}{c|}{2.46}  & 2.49 \\
                                   & Average words per absent keyphrase          & 2.29   & \multicolumn{1}{c|}{2.14} & \multicolumn{1}{c|}{2.62}  & 2.91 \\
                                   & Average no. of present keyphrases per input & 3.28   & \multicolumn{1}{c|}{8.75} & \multicolumn{1}{c|}{9.63}  & 8.29 \\
                                   & Average no. of absent keyphrases per input  & 2.57   & \multicolumn{1}{c|}{2.75} & \multicolumn{1}{c|}{3.46}  & 5.57 \\ \hline
\multirow{5}{*}{\textbf{Semeval}}  & Average words in Title + Abstract           & 183.48 & \multicolumn{1}{c|}{}     & \multicolumn{1}{c|}{}      &      \\
                                   & Average words per present keyphrase         & 1.91   & \multicolumn{1}{c|}{2.02} & \multicolumn{1}{c|}{2.29}  & 2.34 \\
                                   & Average words per absent keyphrase          & 2.22   & \multicolumn{1}{c|}{2.08} & \multicolumn{1}{c|}{2.54}  & 3.42 \\
                                   & Average no. of present keyphrases per input & 6.01   & \multicolumn{1}{c|}{9.83} & \multicolumn{1}{c|}{8.44}  & 7.94 \\
                                   & Average no. of absent keyphrases per input  & 8.53   & \multicolumn{1}{c|}{3.71} & \multicolumn{1}{c|}{3.05}  & 5.96 \\ \hline
\multirow{5}{*}{\textbf{KP20K}}    & Average words in Title + Abstract           & 157.94 & \multicolumn{1}{c|}{}     & \multicolumn{1}{c|}{}      &      \\
                                   & Average words per present keyphrase         & 1.76   & \multicolumn{1}{c|}{2.07} & \multicolumn{1}{c|}{2.37}  & 2.46 \\
                                   & Average words per absent keyphrase          & 2.24   & \multicolumn{1}{c|}{2.18} & \multicolumn{1}{c|}{2.64}  & 3.27 \\
                                   & Average no. of present keyphrases per input & 3.28   & \multicolumn{1}{c|}{9.03} & \multicolumn{1}{c|}{10.11} & 8.76 \\
                                   & Average no. of absent keyphrases per input  & 2.01   & \multicolumn{1}{c|}{2.41} & \multicolumn{1}{c|}{3.54}  & 5.79 \\ \hline
\multirow{5}{*}{\textbf{KPTimes}}  & Average words in Title + Abstract           & 643.24 & \multicolumn{1}{c|}{}     & \multicolumn{1}{c|}{}      &      \\
                                   & Average words per present keyphrase         & 1.48   & \multicolumn{1}{c|}{1.75} & \multicolumn{1}{c|}{2.44}  & 2.23 \\
                                   & Average words per absent keyphrase          & 2.36   & \multicolumn{1}{c|}{2.05} & \multicolumn{1}{c|}{3.01}  & 2.83 \\
                                   & Average no. of present keyphrases per input & 3.18   & \multicolumn{1}{c|}{9.84} & \multicolumn{1}{c|}{9.95}  & 9.82 \\
                                   & Average no. of absent keyphrases per input  & 1.92   & \multicolumn{1}{c|}{2.27} & \multicolumn{1}{c|}{8.4}   & 6.74 \\ \hline
\end{tabular}}
\caption{Statistics of datasets and the model generations.} 
\label{tb:stats}
\end{table}

\section{Qualitative Analysis of Keyphrase Generation}
\label{sec:qualitative_analyses}

In the results presented in the main paper, we find that the LLMs perform quite well in Inspec, quite poorly in KPTimes, and moderately competitively in the other datasets. Our analyses, here, provide some insights about why this happens. As the anecdotal examples in Table \ref{tab:example_gen} and Table \ref{tab:example2} show, the LLMs (particularly GPT-4o) are biased towards generating high number of keyphrases ($\sim 10$) and more of multi-word keyphrases. Moreover, they are biased towards generating more present keyphrases than absent. This pattern of generation matches very well with the pattern of annotated keyphrases in Inspec (larger number of keyphrases and bigger multi-word keyphrases). On the other hand, the annotated keyphrases in KPTimes are on the opposite side of the spectrum. They have fewer keyphrases compared to other datasets, and short (typically single word) keyphrases. The statistics of other datasets are in the middle of the spectrum. These trends that we observe in a few anecdotal examples, are also backed quantitatively in Table \ref{tb:stats}. The table shows several statistics like average number of words per present or absent keyphrases and average number of present and absent keyphrases per input both in model generations and the dataset ground truths. As can be seen, the statistics of model generations correspond most closely to Inspec and least closely to KPTimes. Moreover, the higher average input size of KPTimes may also make things harder for the LLMs. SemEval also has ground truths with higher number of keyphrases comparable to Inspec, but it has much higher ratios of absent keyphrases which conflicts with the pattern of model generations.  

All these points can provide a few insights as to why the LLMs perform best in Inspec, worst in KPTimes and neither very good nor very bad in the other datasets.

\begin{table*}[t]
\centering
\scriptsize
\def\arraystretch{1.2}
\resizebox{0.9\textwidth}{!}{%
\begin{tabular}{l|cc|cc|cc|cc|cc}
\hline
 & \multicolumn{2}{c}{\textbf{Inspec}} & \multicolumn{2}{|c}{\textbf{Krapivin}} & \multicolumn{2}{|c}{\textbf{SemEval}} & \multicolumn{2}{|c}{\textbf{KP20K}} & \multicolumn{2}{|c}{\textbf{KPTimes}} \\  
\multicolumn{1}{c|}{\textbf{Models}} & \textbf{F1@M} & {\textbf{F1@5}}  & \textbf{F1@M} & {\textbf{F1@5}} & \textbf{F1@M} & {\textbf{F1@5}} & \textbf{F1@M} & {\textbf{F1@5}} & \textbf{F1@M} & \textbf{F1@5} \\ \hline
\multicolumn{11}{c}{\textbf{Present Keyphrase Generation}}\\ \hline
\multicolumn{11}{l}{\textbf{Llama-3.0 8B Instruct}}\\ \hline
Baseline & \textbf{48.3} & 40.5 & 30.9 & \textbf{32.4} & 35.5 & 36.2 & 27.7 & \textbf{30.7} & 27.0 & \textbf{31.3}\\
\hdashline
\multicolumn{11}{l}{Beam Search  (Beam width=10)} \\
\hdashline
Union & 38.1& 46.7& 24.2  & 27.3  & 27.3 & 33.1 & 20.3& 23.5& 14.9 & 18.8 \\
Union Concat & 44.4& \textbf{52.0}& \textbf{32.5}  & 30.7  & \textbf{36.4} & 37.0 & 30.7& 27.2& 31.2 & 22.9\\
Union Interleaf & 41.5& 49.4& \textbf{32.5}  & 31.1  & \textbf{36.4} & \textbf{37.2} & \textbf{31.1}& 28.0& \textbf{31.7} & 23.6\\
Frequency Order & 46.3 & 52.2 & 31.7 & 31.1 & 34.8 & 37.0 & 29.5 & 27.4 & 25.0 & 22.9\\
\hline

\multicolumn{11}{l}{\textbf{Phi-3.0 3.8B Mini 128K Instruct}}\\ \hline
Baseline & \textbf{48.2} & 42.2 & 22.2 & 22.5 & 28.4 & 28.6 & 17.6 & \textbf{19.1} & 9.3 & \textbf{11.2}\\
\hdashline
\multicolumn{11}{l}{Beam Search  (Beam width=10)} \\
\hdashline
Union & 36.5 & 46.4 & 16.7 & 20.4 & 17.3 & 24.9 & 12.5 & 14.8 & 4.8 & 6.4\\
Union Concat & 45.1 & 52.1 & 22.1 & 22.2 & 28.9 & 28.9 & 19.0 & 16.7 & 10.3 & 7.3\\
Union Interleaf & 44.4 & 51.0 & \textbf{22.5} & \textbf{22.6} & \textbf{29.8} & \textbf{29.7} & \textbf{19.7} & 17.5 & \textbf{10.4} & 7.5\\
Frequency Order & 45.0 & \textbf{52.4} & 21.7 & 22.1 & 25.2 & 28.7 & 16.6 & 16.7 & 5.8 & 7.2\\

\hline
\multicolumn{11}{c}{\textbf{Absent Keyphrase Generation}}\\ \hline
\multicolumn{11}{l}{\textbf{Llama-3.0 8B Instruct}}\\ \hline
Baseline & 6.8 & 5.5 & 4.6 & 3.8 & \textbf{3.2} & 3.0 & 3.8 & 3.0 & 4.6 & 3.6\\
\hdashline
\multicolumn{11}{l}{Beam Search  (Beam width=10)} \\
\hdashline
Union  & 7.9 & 6.2 & 4.3 & 3.9 & 2.7 & 2.9 & 4.4 & 3.8 & 4.4 & 3.9\\
Union Concat & 8.1 & \textbf{8.0} & 4.8 & 4.5 & 2.8 & \textbf{3.2} & \textbf{4.5} & 4.3 & \textbf{4.7} & 4.7\\
Union Interleaf & 8.1 & 6.9 & 4.7 & 4.3 & 2.6 & 2.2 & \textbf{4.5} & \textbf{4.4} & \textbf{4.7} & \textbf{4.8} \\
Frequency Order  & \textbf{8.2} & 7.7 & \textbf{4.9} & \textbf{4.6} & 2.8 & \textbf{3.2} & \textbf{4.5} & 4.3 & \textbf{4.7} & \textbf{4.8}\\
\hline
\multicolumn{11}{l}{\textbf{Phi-3.0 3.8B Mini 128K Instruct}}\\ \hline
Baseline & 7.3 & 6.3 & 1.3 & 1.1 & \textbf{2.0} & \textbf{1.5} & 1.3 & 1.1 & 0.4 & 0.4\\
\hdashline
\multicolumn{11}{l}{Beam Search  (Beam width=10)} \\
\hdashline
Union & 8.3 & 7.8 & 1.6 & 1.3 & 0.8 & 0.7 & 1.4 & 1.2 & 0.4 & 0.3\\
Union Concat & 9.0 & \textbf{9.5} & \textbf{1.8} & 1.5 & 1.4 & 1.4 & \textbf{1.5} & \textbf{1.5} & 0.4 & 0.4\\
Union Interleaf & 9.0 & 8.5 & 1.7 & \textbf{1.6} & 1.4 & \textbf{1.5} & \textbf{1.5} & \textbf{1.5} & 0.4 & 0.4\\
Frequency Order & \textbf{9.3} & \textbf{9.5} & 1.7 & 1.5 & 1.4 & 1.3 & \textbf{1.5} & \textbf{1.5} & \textbf{0.5} & \textbf{0.5}\\
\hline
\end{tabular}}
\caption{Comparison of baseline models and beam search models with different aggregation strategies for both present and absent keyphrase generation.} 
\label{tb:beamsearch}
\end{table*}


\begin{table*}[t]
\centering
\scriptsize
\def\arraystretch{1.2}
\resizebox{0.9\textwidth}{!}{%
\begin{tabular}{l|cc|cc||cc|cc}
\hline
 & \multicolumn{4}{c||}{\textbf{Present Keyphrase Generation}} & \multicolumn{4}{c}{\textbf{Absent Keyphrase Generation}} \\  
\multicolumn{1}{c|}{\textbf{Models}} & \multicolumn{2}{c|}{\textbf{KP20K}} & \multicolumn{2}{c||}{\textbf{KPTimes}} & \multicolumn{2}{c|}{\textbf{KP20K}} & \multicolumn{2}{c}{\textbf{KPTimes}} \\  
\cline{2-9} 
 & \textbf{F1@M} & \textbf{F1@5} & \textbf{F1@M} & \textbf{F1@5} & \textbf{F1@M} & \textbf{F1@5} & \textbf{F1@M} & \textbf{F1@5} \\ 
\hline

\multicolumn{9}{c}{\textbf{Llama-3.0 8B Instruct}} \\ 
\hline
Baseline & 26.8 & 30.0 & \textbf{28.3} & 33.0 & 3.2 & 3.0 & \textbf{3.9} & \textbf{3.8} \\ 
\hdashline
Union & 18.0 & 15.1 & 13.2 & 9.4 & 2.7 & 3.3 & 1.3 & 1.6 \\ 
Union Concat & 26.8 & 30.3 & 26.1 & 33.2 & 4.8 & 4.5 & 3.1 & 3.0 \\ 
Union Interleaf &\textbf{28.1} & 30.4 & 28.2 & \textbf{33.4 }& 5.0 & 4.7 & 3.1 & 3.0 \\ 
Frequency Order & 27.4 & \textbf{30.6} & 26.3 & 31.9 & \textbf{5.5 }& \textbf{4.9} & 3.5 & 3.2 \\ 
\hline

\multicolumn{9}{c}{\textbf{Phi-3.0 3.8B Mini 128K Instruct}} \\ 
\hline
Baseline & 17.2 & 19.2 & 9.7 & 11.6 & 1.2 & 1.2 & \textbf{0.4} & \textbf{0.4} \\ 
\hdashline
Union & 12.3 & 10.6 & 6.2 & 4.8 & 0.7 & 0.7 & 0.2 & 0.2 \\ 
Union Concat & 18.8 & 20.6 & 11.7 & 12.5 & 1.6 & 1.7 & 0.3 & \textbf{0.4} \\ 
Union Interleaf & \textbf{21.0} & \textbf{22.0} & \textbf{15.8} & \textbf{14.9} & 1.6 & 1.5 & 0.3 & \textbf{0.4} \\ 
Frequency Order & 19.2 & 19.6 & 11.0 & 10.5 & \textbf{1.9} & \textbf{ 2.1} & \textbf{0.4} & \textbf{0.4} \\ 
\hline

\multicolumn{9}{c}{\textbf{GPT-4o}} \\ 
\hline
Baseline & 20.1 & 24.7 & 11.4 & 14.7 & 2.4 & 2.5 & 0.4 & 0.5 \\ 
\hdashline
Union & 15.7 & 13.2 & 8.8 & 7.4 & 1.3 & 1.5 & 0.3 & 0.2 \\ 
Union Concat & 20.1 & 24.6 & 12.9 & 15.5 & 2.5 & 2.3 & 0.4 & 0.5 \\ 
Union Interleaf & \textbf{21.9} & \textbf{26.3} & \textbf{16.0} & \textbf{18.2} & \textbf{2.8} & \textbf{2.8} & \textbf{0.6} & \textbf{0.7} \\ 
Frequency Order & 20.0 & 21.7 & 12.1 & 12.4 & 2.6 & 2.6 & \textbf{0.6} & 0.6 \\ 
\hline
\end{tabular}}
\caption{Comparison of baseline models and multi-sampling models on a subsample of 2,000, using different aggregation strategies for both present and absent keyphrase generation.}
\label{tb:multisampling_2000}
\end{table*}

\begin{table*}[]
\def\arraystretch{1.2}
\resizebox{\textwidth}{!}{%
\centering
\begin{tabular}{|c|c|c|c|c|c|}
\hline
\textbf{Dataset} &
  \textbf{Inspec} &
  \textbf{Krapivin} &
  \textbf{SemEval} &
  \textbf{KP20K} &
  \textbf{KPTimes} \\ \hline
\textbf{Title} &
  \multicolumn{1}{c|}{\begin{tabular}[c]{@{}c@{}}Loudspeaker Voice-Coil Inductance Losses:\\ Circuit Models, Parameter Estimation, and\\ Effect on Frequency Response\end{tabular}} &
  \multicolumn{1}{c|}{\begin{tabular}[c]{@{}c@{}}computation in networks of\\  passively mobile finite state\\  sensors\end{tabular}} &
  \multicolumn{1}{c|}{\begin{tabular}[c]{@{}c@{}}Computing the Banzhaf Power Index \\ in Network Flow Games\end{tabular}} &
  \multicolumn{1}{c|}{\begin{tabular}[c]{@{}c@{}}A Graph Coloring Based TDMA \\ Scheduling Algorithm for Wireless \\ Sensor Networks.\end{tabular}} &
  \begin{tabular}[c]{@{}c@{}}Auto sales slide 7.6\% in May on \\ minicar tax.\end{tabular} \\ \hline
\textbf{Abstract} &
  \begin{tabular}[c]{@{}c@{}} When the series resistance is\\ separated and treated as a separate\\ element, it is shown that losses in an\\ inductor require the ratio of the flux to\\ MMF in the core to be frequency\\ dependent. For small-signal operation,\\ this dependence leads to a circuit model\\ composed of a lossless inductor and a\\ resistor in parallel, both of which are\\ frequency dependent. Mathematical\\ expressions for these elements are\\ derived under the assumption that the\\ ratio of core flux to MMF varies as\\ $\omega^{n-1}$, where n is a constant.\\ A linear regression technique is\\ described for extracting the model\\ parameters from measured data.\\ Experimental data are presented to\\ justify the model for the lossy inductance\\ of a loudspeaker voice-coil. A SPICE\\ example is presented to illustrate the\\ effects of voice-coil inductor losses on\\ the frequency response of a typical\\ driver\end{tabular} &
  
  \begin{tabular}[c]{@{}c@{}} we explore\\ the computational power of networks of\\ small resource limited mobile agents .\\ we define two new models of\\ computation based on pairwise\\ interactions of finite state agents in\\ populations of finite but unbounded size\\ . with a fairness condition on interactions\\ , we define the concept of stable\\ computation of a function or predicate ,\\ and give protocols that stably compute\\ functions in a class including boolean\\ combinations of threshold k , parity ,\\ majority , and simple arithmetic . we\\ prove that all stably computable\\ predicates are in nl . with uniform\\ random sampling of pairs to interact , we\\ define the model of conjugating\\ automata and show that any counter\\ machine with o (n) counters of capacity o\\ ( n ) can be simulated with high\\ probability by a protocol in a population\\ of size n ...\end{tabular} &
  
  \begin{tabular}[c]{@{}c@{}} Preference\\ aggregation is used in a variety of\\ multiagent applications, and as a result,\\ voting theory has become an important topic\\ in multiagent system research. However,\\ power indices (which reflect how much real\\ power a voter has in a weighted voting\\ system) have received relatively little\\ attention, although they have long been\\ studied in political science and economics.\\ The Banzhaf power index is one of the most\\ popular; it is also well-defined for any\\ simple coalitional game. In this paper, we\\ examine the computational complexity of\\ calculating the Banzhaf power index within\\ a particular multiagent domain, a network\\ flow game. Agents control the edges of a\\ graph; a coalition wins if it can send a flow\\ of a given size from a source vertex to a\\ target vertex. The relative power of each\\ edge/agent reflects its significance in\\ enabling such a flow, and in real-world\\ networks could be used, for example, to\\ allocate resources for maintaining parts of\\ the network... \end{tabular} &
  
  \begin{tabular}[c]{@{}c@{}} Wireless sensor\\ networks should provide with valuable\\ service, which is called service-oriented\\ requirement. To meet this need, a novel\\ distributed graph coloring based time\\ division multiple access scheduling\\ algorithm (GCSA), considering real-time\\ performance for clustering-based sensor\\ network, is proposed in this paper, to\\ determine the smallest length of\\ conflict-free assignment of timeslots for\\ intra-cluster transmissions. GCSA\\ involves two phases. In coloring phase,\\ networks are modeled using graph\\ theory, and a distributed vertex coloring\\ algorithm, which is a distance-2 coloring\\ algorithm and can get colors near to\\ $\delta +1$, is proposed to assign a\\ color to each node in the network. Then,\\ in scheduling phase, each independent\\ set is mapped to a unique timeslot\\ according to the sets priority which is\\ obtained by considering network\\ structure... \end{tabular} &
  
  \begin{tabular}[c]{@{}c@{}} Auto sales in May fell 7.6 percent to\\ 335,644 units from a year ago as the\\ April tax hike on minivehicles weighed\\ on demand, industry bodies said\\ Monday. Minicar sales sank 19.6\\ percent to 125,755 units, down for the\\ fifth consecutive month, the Japan Light\\ Motor Vehicle and Motorcycle\\ Association said. Minivehicles, which\\ have engine displacements no larger\\ than 660cc, account for around 40\\ percent of new car sales in Japan. Sales\\ of other cars meanwhile rose 1.4\\ percent to 209,889 units, rising for the\\ second consecutive month, the Japan\\ Automobile Dealers Association said,\\ hinting the impact of the April 2014\\ consumption tax hike is on the wane.\\ Demand for cars was sluggish during\\ the fiscal year ended March 31 after the\\ first stage of the doubling of the\\ consumption tax raised the levy by 3\\ points to 8 percent, tipping Japan into\\ yet another recession. It was the\\ nation’s first tax hike in 17 years. The\\ second stage, which has been delayed,\\ will raise it to 10 percent.
  \end{tabular} \\ \hline
\textbf{\begin{tabular}[c]{@{}c@{}}Ground Truth \\ Keyphrases\end{tabular}} &
  \multicolumn{1}{c|}{\begin{tabular}[c]{@{}c@{}}{[}"\textcolor{blue}{loudspeaker voice-coil inductance losses}", \\ "\textcolor{blue}{circuit models}", "\textcolor{blue}{parameter estimation}", \\ "\textcolor{blue}{frequency response}", "\textcolor{blue}{series resistance}", \\ "\textcolor{blue}{small-signal operation}", "lossless inductor", \\ "\textcolor{blue}{linear regression}", "\textcolor{blue}{lossy inductance}", \\ "\textcolor{blue}{SPICE}", "loudspeaker driver", \\ "core flux to MMF ratio"{]}\end{tabular}} &
  \multicolumn{1}{c|}{\begin{tabular}[c]{@{}c@{}}{[}"\textcolor{blue}{mobile agent}",\\  "\textcolor{blue}{finite state agent}", \\ "\textcolor{blue}{stable computation}", \\ "sensor net", \\ "diffuse computation", \\ "intermittent communication"{]}\end{tabular}} &
  \multicolumn{1}{c|}{\begin{tabular}[c]{@{}c@{}}{[}"\textcolor{blue}{banzhaf power index}", "\textcolor{blue}{power index}", \\ "\textcolor{blue}{network flow game}", "\textcolor{blue}{prefer aggreg}", \\ "\textcolor{blue}{multiag applic}", "vote theori", \\ "\textcolor{blue}{vote}", "\textcolor{blue}{comput complex}", \\ "algorithm and problem complex analysi", \\ "social choic theori", "autom agent vote", \\ "probabilist model", "connect game"{]}\end{tabular}} &
  \multicolumn{1}{c|}{\begin{tabular}[c]{@{}c@{}}{[}"\textcolor{blue}{graph coloring}", \\ "\textcolor{blue}{tdma}", \\ "\textcolor{blue}{distributed}", \\ "\textcolor{blue}{real-time}", \\ "clustering"{]}\end{tabular}} &
  \begin{tabular}[c]{@{}c@{}}{[}"autos", \\ "minicars", \\ "\textcolor{blue}{car sales}"{]}\end{tabular} \\ \hline

  \textbf{Llama-3} &
  \begin{tabular}[c]{@{}c@{}}
    {[}`Lossy inductor", \\ 
    `\textcolor{blue}{Loudspeaker voice-coil}", \\
    `\textcolor{blue}{Inductance losses}", \\
    `\textcolor{blue}{Circuit modeling}", \\
    `\textcolor{blue}{Frequency response}", \\
    `\textcolor{blue}{SPICE}", \\ 
    `Magnetic core", `\textcolor{blue}{MMF}", `\textcolor{blue}{Flux}", \\
    `\textcolor{blue}{Linear regression}", \\
    `\textcolor{blue}{Parameter estimation}"]
  \end{tabular} &
  
  \begin{tabular}[c]{@{}c@{}}
    {[}`\textcolor{blue}{finite state sensors}", \\
    `passive mobile agents", \\
    `\textcolor{blue}{computational power}", `\textcolor{blue}{networks}", \\
    `\textcolor{blue}{pairwise interactions}", \\
    `\textcolor{blue}{stable computation}", \\
    `\textcolor{blue}{boolean combinations}", \\
    `\textcolor{blue}{threshold k}", `\textcolor{blue}{parity}", \\ 
    `\textcolor{blue}{majority}", `\textcolor{blue}{arithmetic}", `\textcolor{blue}{NL}", \\
    `\textcolor{blue}{uniform random sampling}", \\
    `\textcolor{blue}{conjugating automata}", \\
    `\textcolor{blue}{counter machine}", \\ 
    `PRAL", `\textcolor{blue}{open problems}", \\ 
    `\textcolor{blue}{future directions}"]
  \end{tabular} &
  
  \begin{tabular}[c]{@{}c@{}}
    {[}`\textcolor{blue}{Computing the Banzhaf Power Index}", \\
    `\textcolor{blue}{Network Flow Games}", \\
    `\textcolor{blue}{Multiagent System}", `\textcolor{blue}{Power Indices}", \\
    `\textcolor{blue}{Preference Aggregation}", `\textcolor{blue}{Voting Theory}", \\
    `Simple Coalitional Games", `\textcolor{blue}{Graph}", \\ 
    `\textcolor{blue}{Network Flow}", `\textcolor{blue}{Coalition}", \\ 
    `Resource Allocation", `\textcolor{blue}{Computational} \\  \textcolor{blue}{Complexity}", `\#P-completeness", \\
    `\textcolor{blue}{Polynomial Algorithm}"]
  \end{tabular} &
  
  \begin{tabular}[c]{@{}c@{}}{[}"\textcolor{blue}{Graph Coloring}", "\textcolor{blue}{TDMA Scheduling}",\\  "\textcolor{blue}{Wireless Sensor Networks}",\\  "\textcolor{blue}{Service-Oriented}",\\  "Distributed Algorithm",\\  "\textcolor{blue}{Real-Time Performance}",\\  "\textcolor{blue}{Conflict-Free Assignment}",\\  "\textcolor{blue}{Time Division Multiple Access}",\\  "\textcolor{blue}{Clustering-Based Sensor Network}", \\ "\textcolor{blue}{Graph Theory}",\\  "\textcolor{blue}{Distributed Vertex Coloring Algorithm}",\\  "\textcolor{blue}{Distance-2 Coloring}",\\ "\textcolor{blue}{Network Structure}",\\  "\textcolor{blue}{Intra-Cluster Delay}", \\ "\textcolor{blue}{Intra-Cluster Throughput}",\\  "\textcolor{blue}{Communication Reliability}"{]}\end{tabular} &
  
  \begin{tabular}[c]{@{}c@{}}{[}"\textcolor{blue}{Auto sales}", "\textcolor{blue}{minicar tax}",\\ "\textcolor{blue}{Japan}", "\textcolor{blue}{April tax hike}",\\ "\textcolor{blue}{minivehicles}", "\textcolor{blue}{car sales}",\\ "\textcolor{blue}{consumption tax}"{]}\end{tabular} \\ \hline

\textbf{Phi-3} &

  \begin{tabular}[c]{@{}c@{}}{[}"Lossy loudspeaker voice-coil", \\ "Inductor inductance losses", \\ "Frequency dependent losses",\\  "Lossless inductor model", \\ "Resistor model",\\  "\textcolor{blue}{Small-signal operation}", "\\ Mathematical expressions derivation",\\  "Core flux to MMF ratio",\\  "\textcolor{blue}{Linear regression technique}", \\ "Loudspeaker frequency response", \\ "\textcolor{blue}{SPICE example}"{]}\end{tabular} &
  
  \begin{tabular}[c]{@{}c@{}}{[}"\textcolor{blue}{small resource limited mobile agents}",\\ "\textcolor{blue}{computation in networks}", \\ "\textcolor{blue}{finite state agents}",\\ "\textcolor{blue}{stable computation}",\\  "\textcolor{blue}{boolean combinations}",\\ "\textcolor{blue}{conjugating automata}",\\  "counter machine simulation",\\  "\textcolor{blue}{high probability}",\\  "\textcolor{blue}{promising future directions}"{]}\end{tabular} &
  
  \begin{tabular}[c]{@{}c@{}}{[}"\textcolor{blue}{Banzhaf power index}",\\  "\textcolor{blue}{network flow game}",\\ "\textcolor{blue}{computational complexity}",\\  "\textcolor{blue}{\#P-complete}",\\ "\textcolor{blue}{polynomial algorithm}",\\  "\textcolor{blue}{coalitional game}",\\  "\textcolor{blue}{voting theory}",\\  "resource allocation",\\  "\textcolor{blue}{real-world networks}",\\ "graph edge significance"{]}\end{tabular} &
  
  \begin{tabular}[c]{@{}c@{}}{[}"\textcolor{blue}{Graph Coloring}",\\  "Time Division Multiple Access - \\ (TDMA) Scheduling", \\ "\textcolor{blue}{Service-Oriented Requirement}", \\ "\textcolor{blue}{Clustering-Based Sensor Network}", \\ "\textcolor{blue}{Distributed Vertex Coloring Algorithm}", \\ "\textcolor{blue}{Distance-2 Coloring Algorithm}", \\ "\textcolor{blue}{Intra-Cluster Transmissions}",\\  "\textcolor{blue}{Intra-Cluster Delay}", \\ "\textcolor{blue}{Intra-Cluster Throughput}", \\ "\textcolor{blue}{Real-Time Performance}",\\  "\textcolor{blue}{Communication Reliability}",\\  "\textcolor{blue}{Experimental Results}"{]}\end{tabular} &
  
  \begin{tabular}[c]{@{}c@{}}{[}"auto sales decline", "minicar tax\\ impact", "minicar sales drop",\\ "\textcolor{blue}{consumption tax hike}", "car\\ demand slump", "Japan"s\\ recession", "tax policy change"{]}\end{tabular} \\ \hline

  \textbf{GPT-4o} &
  \begin{tabular}[c]{@{}c@{}}{[}"\textcolor{blue}{loudspeaker voice-coil inductance}",\\ "\textcolor{blue}{circuit models}", "\textcolor{blue}{parameter estimation}",\\ "\textcolor{blue}{frequency response}", "\textcolor{blue}{inductor losses}",\\ "\textcolor{blue}{series resistance}", "\textcolor{blue}{small-signal} \\ \textcolor{blue}{operation}", "\textcolor{blue}{lossless inductor}", "\textcolor{blue}{resistor} \\ \textcolor{blue}{ in parallel}", "frequency dependent\\ elements", "core flux to MMF ratio",\\ "\textcolor{blue}{mathematical expressions}", "\textcolor{blue}{linear} \\ \textcolor{blue}{regression technique}", "\textcolor{blue}{lossy}\\ \textcolor{blue}{inductance}", "\textcolor{blue}{SPICE example}",\\ "loudspeaker driver"{]}\end{tabular} &
  
  \begin{tabular}[c]{@{}c@{}}{[}"\textcolor{blue}{passively mobile finite state sensors}",\\ "\textcolor{blue}{computational power of networks}",\\  "\textcolor{blue}{small resource limited mobile agents}",\\  "\textcolor{blue}{pairwise interactions}",\\ "\textcolor{blue}{stable computation}", \\ "\textcolor{blue}{boolean combinations}",\\ "threshold functions", "\textcolor{blue}{parity}", \\ "\textcolor{blue}{majority}", "\textcolor{blue}{simple arithmetic}",\\  "\textcolor{blue}{stably computable predicates}",\\  "NL complexity class", \\ "\textcolor{blue}{uniform random sampling}",\\ "\textcolor{blue}{conjugating automata}", \\ "counter machine simulation", \\ "high probability computational model",\\ "PRL complexity class", "\textcolor{blue}{open problems}",\\  "\textcolor{blue}{future directions}"{]}\end{tabular} &
  
  \begin{tabular}[c]{@{}c@{}}{[}"\textcolor{blue}{Banzhaf power index}", \\ "\textcolor{blue}{network flow games}",\\ "multiagent systems", \\ "\textcolor{blue}{computational complexity}",\\ "\textcolor{blue}{power indices}",\\  "\textcolor{blue}{weighted voting system}",\\ "\textcolor{blue}{coalitional game}",\\  "\textcolor{blue}{network flow}", "\textcolor{blue}{\#P-complete}",\\ "polynomial algorithm",\\  "\textcolor{blue}{preference aggregation}",\\ "voting theories", "resource allocation", \\ "\textcolor{blue}{multiagent domain}", \\ "graph edges control"{]}\end{tabular} &
  
  \begin{tabular}[c]{@{}c@{}}{[}"\textcolor{blue}{graph coloring}",\\  "\textcolor{blue}{TDMA scheduling}", \\ "\textcolor{blue}{wireless sensor networks}", \\ "distributed algorithm", \\ "time division access", \\ "\textcolor{blue}{intra-cluster transmissions}", \\ "\textcolor{blue}{clustering-based sensor network}",\\  "\textcolor{blue}{real-time performance}", \\ "\textcolor{blue}{conflict-free assignment}", \\ "\textcolor{blue}{vertex coloring algorithm}", \\ "\textcolor{blue}{independent set}", "\textcolor{blue}{intra-cluster delay}",\\  "\textcolor{blue}{intra-cluster throughput}",\\  "\textcolor{blue}{communication reliability}"{]}\end{tabular} &
  
  \begin{tabular}[c]{@{}c@{}}{[}"auto sales decline", "minicar tax\\ impact", "Japan auto market",\\ "minivehicle sales", "\textcolor{blue}{consumption }\\ \textcolor{blue}{tax hike}", "Japan recession",\\ "minicar sales decrease", "tax\\ impact on demand", "car sales\\ increase", "\textcolor{blue}{Japan Light Motor} \\ \textcolor{blue}{Vehicle and Motorcycle}\\ \textcolor{blue}{Association}", "\textcolor{blue}{Japan Automobile}\\ \textcolor{blue}{Dealers Association}", "fiscal year\\ car sales", "tax hike effects"{]}\end{tabular} \\ \hline

\end{tabular}}
\caption{An example of input document from each dataset, along with the corresponding keyphrase generations by various models, where blue represents present keyphrases and black represents absent keyphrases.}
\label{tab:example_gen}
\end{table*}

\begin{table*}[]
\def\arraystretch{1.2}
\resizebox{\textwidth}{!}{%
\centering
\begin{tabular}{|c|c:c|c:c|}
\hline
\textbf{Dataset} &
  \multicolumn{2}{c|}{\textbf{Inspec}} &
  \multicolumn{2}{c|}{\textbf{KPTimes}} \\ \hline
\textbf{Title} &
  \multicolumn{1}{c:}{\begin{tabular}[c]{@{}c@{}}WEXTOR: a Web-based tool for generating\\  and visualizing experimental designs and \\ procedures\end{tabular}} & 
  \begin{tabular}[c]{@{}c@{}}A framework for evaluating the \\ data-hiding capacity of image \\ sources\end{tabular} &
  \multicolumn{1}{c:}{\begin{tabular}[c]{@{}c@{}}Chinese tourists step up for Abe as Japanese \\ tighten belts\end{tabular}} &
  \begin{tabular}[c]{@{}c@{}}Manafort family business defends name\\  as cousin sits in jail\end{tabular} \\ \hline
\textbf{Abstract} &
  \multicolumn{1}{c:}{\begin{tabular}[c]{@{}c@{}}WEXTOR is a Javascript-based experiment\\ generator and teaching tool on the World Wide\\ Web that can be used to design laboratory\\ and Web experiments in a guided step-by-step\\ process. It dynamically creates the customized\\ Web pages and Javascripts needed for the\\ experimental procedure and provides\\ experimenters with a print-ready visual display\\ of their experimental design. WEXTOR flexibly\\ supports complete and incomplete factorial\\ designs with between-subjects,\\ within-subjects, and quasi-experimental\\ factors, as well as mixed designs. The\\ software implements client-side response time\\ measurement and contains a content wizard\\ for creating interactive materials, as well as\\ dependent measures (graphical scales,\\ multiple-choice items, etc.), on the experiment\\ pages...\end{tabular}} &
  \begin{tabular}[c]{@{}c@{}}An information-theoretic model for image\\ watermarking and data hiding is presented in\\ this paper. Previous theoretical results are\\ used to characterize the fundamental capacity\\ limits of image watermarking and data-hiding\\ systems. Capacity is determined by the\\ statistical model used for the host image, by\\ the distortion constraints on the data hider and\\ the attacker, and by the information available\\ to the data hider, to the attacker, and to the\\ decoder. We consider autoregressive,\\ block-DCT, and wavelet statistical models for\\ images and compute data-hiding capacity for\\ compressed and uncompressed host-image\\ sources.\end{tabular} &
  \multicolumn{1}{c:}{\begin{tabular}[c]{@{}c@{}}When Jingyan Hou made her first trip to Japan\\ in 1997, the office worker from Beijing spent\\ ¥200,000 during a weeklong stay on\\ accommodations, meals, transport and\\ souvenirs. On her second visit this year, she\\ spent that much on just one Louis Vuitton\\ handbag in Tokyo’s Ginza shopping district.\\ The increasing wealth of travelers like Hou,\\ 45, underscores the opportunity for Japan to\\ expand its tourism industry as China’s\\ burgeoning middle class goes on vacations\\ abroad. The yen’s slump to a seven-year low\\ against the dollar is also broadening the\\ country’s appeal globally and bolstering the\\ Abe administration’s effort to double visitors by\\ the 2020 Tokyo Olympics. “There’s a lot of\\ room to boost the number of foreign tourists\\ coming to Japan with these growing\\ economies in our neighborhood,” said Daiki\\ Takahashi, an economist at the Dai-ichi Life\\ Research Institute in Tokyo...\end{tabular}} &
  \begin{tabular}[c]{@{}c@{}}What do you do if you share a name with one\\ of the most prominent defendants in the\\ special counsel’s investigation into Russia?\\ Paul Manafort’s daughter decided to change\\ her name. Leaders of New Britain,\\ Connecticut, considered renaming Paul\\ Manafort Drive, a street named after his father.\\ At Manafort Brothers Inc., a family-owned New\\ England construction firm, they are defending\\ the Manafort name and legacy while\\ distancing themselves from their cousin,\\ Trump’s former campaign chairman who was\\ recently blasted by prosecutors for years of\\ lies and lawbreaking. The Manafort name has\\ been a familiar one in New England politics\\ and business for decades, creating a\\ predicament for the family as the 69-year-old\\ former attorney is scheduled to be sentenced\\ Thursday. Manafort Brothers is one of New\\ England’s best known construction\\ companies...\end{tabular} \\ \hline
\textbf{\begin{tabular}[c]{@{}c@{}}Ground Truth \\ Keyphrases\end{tabular}} &
  \multicolumn{1}{c:}{\begin{tabular}[c]{@{}c@{}}{[}'\textcolor{blue}{WEXTOR}', '\textcolor{blue}{Web-based tool}',\\ '\textcolor{blue}{Javascript-based experiment generator}',\\ '\textcolor{blue}{teaching tool}', '\textcolor{blue}{World Wide Web}', '\textcolor{blue}{customized}\\ \textcolor{blue}{Web pages}', '\textcolor{blue}{print-ready visual display}',\\ '\textcolor{blue}{factorial designs}', '\textcolor{blue}{client-side response time}\\ \textcolor{blue}{measurement}', '\textcolor{blue}{content wizard}', '\textcolor{blue}{HTML}', '\textcolor{blue}{Web}\\ \textcolor{blue}{server}', '\textcolor{blue}{logfiles}', '\textcolor{blue}{database}', 'experimental\\ design visualization', 'free software'{]}\end{tabular}} &
  \begin{tabular}[c]{@{}c@{}}{[}'\textcolor{blue}{data-hiding capacity}', '\textcolor{blue}{image sources}',\\ '\textcolor{blue}{information-theoretic model}', '\textcolor{blue}{watermarking}',\\ '\textcolor{blue}{capacity limits}', '\textcolor{blue}{statistical model}', '\textcolor{blue}{distortion}\\ \textcolor{blue}{constraints}', '\textcolor{blue}{wavelet statistical models}',\\ '\textcolor{blue}{uncompressed host-image sources}',\\ '\textcolor{blue}{closed-form expressions}', '\textcolor{blue}{sparse-model}\\ \textcolor{blue}{approximations}', '\textcolor{blue}{geometric attacks}', '\textcolor{blue}{distortion}\\ \textcolor{blue}{measures}', 'autoregressive statistical models',\\ 'block-DCT statistical models', '\textcolor{blue}{compressed}\\ \textcolor{blue}{host-image sources}'\end{tabular} &
  \multicolumn{1}{c:}{\begin{tabular}[c]{@{}c@{}}{[}'\textcolor{blue}{tourism}', '\textcolor{blue}{china}', 'tokyo 2020', \\ '2020 olympics', 'inbound tourism'{]}\end{tabular}} &
  \begin{tabular}[c]{@{}c@{}}{[}'\textcolor{blue}{paul manafort}',\\  '\textcolor{blue}{manafort brothers}',\\  '\textcolor{blue}{new england}',\\  'robert mueller',\\  'donald trump',\\  'russia probe'{]}\end{tabular} \\ \hline

\textbf{Llama-3} &
  \multicolumn{1}{c:}{\begin{tabular}[c]{@{}c@{}}{[}'Web-based experiment generator',\\ '\textcolor{blue}{experimental design}', '\textcolor{blue}{factorial design}',\\ '\textcolor{blue}{response time measurement}', '\textcolor{blue}{client-side}',\\ '\textcolor{blue}{HTML editor}', '\textcolor{blue}{Web experimental design}',\\ 'educational purposes', '\textcolor{blue}{noncommercial}\\ \textcolor{blue}{purposes}', 'psycholinguistics'{]}\end{tabular}} &
  \begin{tabular}[c]{@{}c@{}}{[}'\textcolor{blue}{Image watermarking}', '\textcolor{blue}{data hiding}',\\ '\textcolor{blue}{information-theoretic model}', '\textcolor{blue}{capacity limits}',\\ '\textcolor{blue}{image sources}', '\textcolor{blue}{statistical models}',\\ '\textcolor{blue}{autoregressive}', '\textcolor{blue}{block-DCT}', '\textcolor{blue}{wavelet}',\\ '\textcolor{blue}{compressed}', '\textcolor{blue}{uncompressed host-image}',\\ '\textcolor{blue}{sparse-model}', '\textcolor{blue}{geometric attacks}', '\textcolor{blue}{distortion}\\ \textcolor{blue}{measures}'{]}\end{tabular} &
  \multicolumn{1}{c:}{\begin{tabular}[c]{@{}c@{}}{[}'\textcolor{blue}{China}', '\textcolor{blue}{Japan}', '\textcolor{blue}{Abe}', '\textcolor{blue}{tourism}', '\textcolor{blue}{economy}',\\ '\textcolor{blue}{Tokyo Olympics}', '\textcolor{blue}{travel}', '\textcolor{blue}{shopping}', '\textcolor{blue}{Louis}\\ \textcolor{blue}{Vuitton}', '\textcolor{blue}{Ginza}', '\textcolor{blue}{Beijing}', '\textcolor{blue}{accommodation}',\\ '\textcolor{blue}{consumption tax}', '\textcolor{blue}{recession}', '\textcolor{blue}{World Tourism}\\ \textcolor{blue}{Organization}', '\textcolor{blue}{South Korea}', '\textcolor{blue}{Taiwan}', '\textcolor{blue}{Tokyo}',\\ '\textcolor{blue}{Mount Fuji}', '\textcolor{blue}{Kyoto}', '\textcolor{blue}{Sakaiminato}', '\textcolor{blue}{Tottori}',\\ '\textcolor{blue}{cruise ship}', '\textcolor{blue}{visa}', '\textcolor{blue}{Okinawa}', '\textcolor{blue}{Indonesia}',\\ '\textcolor{blue}{Philippines}', '\textcolor{blue}{Vietnam}', '\textcolor{blue}{customer service}',\\ '\textcolor{blue}{transport}', '\textcolor{blue}{cultural attraction}', '\textcolor{blue}{hostel}', '\textcolor{blue}{love}\\ \textcolor{blue}{hotel}', '\textcolor{blue}{Asakusa}'{]}\end{tabular}} &
  \begin{tabular}[c]{@{}c@{}}{[}'\textcolor{blue}{Paul Manafort}', '\textcolor{blue}{Manafort Brothers}',\\ '\textcolor{blue}{construction company}', '\textcolor{blue}{family business}',\\ 'Russia investigation', '\textcolor{blue}{special counsel}',\\ '\textcolor{blue}{Russia}', '\textcolor{blue}{politics}', '\textcolor{blue}{business}', '\textcolor{blue}{Connecticut}',\\ '\textcolor{blue}{New England}', '\textcolor{blue}{Trump}', '\textcolor{blue}{campaign chairman}',\\ 'lawyer', '\textcolor{blue}{jail}', '\textcolor{blue}{conviction}', '\textcolor{blue}{sentencing}', '\textcolor{blue}{fraud}',\\ '\textcolor{blue}{politics}', '\textcolor{blue}{family name}', '\textcolor{blue}{legacy}', '\textcolor{blue}{reputation}',\\ '\textcolor{blue}{morals}', '\textcolor{blue}{ethics}', 'philanthropy'{]}\end{tabular} \\ \hline
\textbf{Phi-3} &
  \multicolumn{1}{c:}{\begin{tabular}[c]{@{}c@{}}'Experimental design generator', 'Web-based\\ experiment tool', 'Guided procedure design',\\ 'Dynamic content creation', 'Flexible factorial\\ design', 'Client-side response measurement',\\ '\textcolor{blue}{Content wizard}', 'Interactive materials design',\\ '\textcolor{blue}{Methodological features}', 'Web server\\ compatibility', 'Educational resource', '\textcolor{blue}{Freely}\\ \textcolor{blue}{available}', '\textcolor{blue}{World Wide Web}', '\textcolor{blue}{GenPSYLab}',\\ 'Login link'{]}\end{tabular}} &
  \begin{tabular}[c]{@{}c@{}}{[}'\textcolor{blue}{Information-theoretic model}', '\textcolor{blue}{Image}\\ \textcolor{blue}{watermarking}', '\textcolor{blue}{Fundamental capacity limits}',\\ '\textcolor{blue}{Statistical model}', '\textcolor{blue}{Distortion constraints}',\\ 'Data hider vs. attacker', 'Compressed vs.\\ uncompressed images', 'Sparse model\\ approximations', '\textcolor{blue}{Geometric attacks}',\\ '\textcolor{blue}{Distortion measures}', 'Invariant attacks'{]}\end{tabular} &
  \multicolumn{1}{c:}{\begin{tabular}[c]{@{}c@{}}{[}'Chinese tourists growth', 'Japanese tourism\\ industry expansion', "Abe administration's\\ doubling tourists", "yen's slump", "Japanese\\ economy's recession", 'foreign visitors\\ spending', 'duty-free sales increase', '\textcolor{blue}{bilingual}\\ \textcolor{blue}{sales assistants}', 'tourism in less-visited\\ areas', "Tottori's tourist infrastructure", 'visa\\ policy changes', 'inexpensive\\ accommodations', 'Manryo hostels', 'foreign\\ tourist increase'{]}\end{tabular}} &
  \begin{tabular}[c]{@{}c@{}}'\textcolor{blue}{Manafort family business}', '\textcolor{blue}{defend name}',\\ '\textcolor{blue}{Paul Manafort}', '\textcolor{blue}{criminal issues}', '\textcolor{blue}{Manafort}\\ \textcolor{blue}{Brothers Inc.}', '\textcolor{blue}{strong moral and business}\\ \textcolor{blue}{ethics}', 'family legacy', "Paul Manafort's\\ arrest", '\textcolor{blue}{Paul Manafort Drive}', "Paul Sr.'s\\ political scandals", "Paul Jr.'s fight to\\ dissociate", 'minority subcontractor scandal',\\ 'Manafort family foundation', "dissociation from\\ name by Manafort's daughter "{]}\end{tabular} \\ \hline

  \textbf{GPT-4o} &
  \multicolumn{1}{c:}{\begin{tabular}[c]{@{}c@{}}{[}'\textcolor{blue}{WEXTOR}', '\textcolor{blue}{web-based tool}', '\textcolor{blue}{experimental}\\ \textcolor{blue}{designs}', '\textcolor{blue}{Javascript-based experiment}\\ \textcolor{blue}{generator}', 'laboratory experiments', '\textcolor{blue}{web}\\ \textcolor{blue}{experiments}', '\textcolor{blue}{factorial designs}',\\ 'between-subjects factors', 'within-subjects\\ factors', '\textcolor{blue}{quasi-experimental factors}', '\textcolor{blue}{mixed}\\ \textcolor{blue}{designs}', '\textcolor{blue}{client-side response time}\\ \textcolor{blue}{measurement}', '\textcolor{blue}{content wizard}', '\textcolor{blue}{interactive}\\ \textcolor{blue}{materials}', '\textcolor{blue}{dependent measures}', '\textcolor{blue}{graphical}\\ \textcolor{blue}{scales}', '\textcolor{blue}{multiple-choice items}', '\textcolor{blue}{web}\\ \textcolor{blue}{experimental design}', '\textcolor{blue}{platform independent}',\\ 'educational purposes', '\textcolor{blue}{noncommercial}\\ \textcolor{blue}{purposes}', 'visualization of experimental\\ design', 'experiment procedures', '\textcolor{blue}{customized}\\ \textcolor{blue}{web pages}'{]}\end{tabular}} &
  
  \begin{tabular}[c]{@{}c@{}}{[}'\textcolor{blue}{data-hiding capacity}', '\textcolor{blue}{image watermarking}',\\ '\textcolor{blue}{information-theoretic model}', '\textcolor{blue}{image sources}',\\ '\textcolor{blue}{statistical model}', '\textcolor{blue}{host image}', '\textcolor{blue}{distortion}\\ \textcolor{blue}{constraints}', '\textcolor{blue}{data hider}', '\textcolor{blue}{attacker}', '\textcolor{blue}{decoder}',\\ 'autoregressive model', 'block-DCT model',\\ 'wavelet model', '\textcolor{blue}{compressed host-image}\\ \textcolor{blue}{sources}', '\textcolor{blue}{uncompressed host-image sources}',\\ '\textcolor{blue}{sparse-model approximations}', '\textcolor{blue}{geometric}\\ \textcolor{blue}{attacks}', '\textcolor{blue}{distortion measures}', '\textcolor{blue}{capacity limits}',\\ '\textcolor{blue}{closed-form expressions}'{]}\end{tabular} &
  \multicolumn{1}{c:}{\begin{tabular}[c]{@{}c@{}}{[}'\textcolor{blue}{Chinese tourists}', 'Japan tourism industry',\\ "Shinzo Abe's growth strategy", "yen's slump",\\ 'Tokyo Olympics 2020', 'consumption tax\\ increase', 'foreign tourist spending', 'inbound\\ tourism', 'consumer spending', '\textcolor{blue}{World Tourism}\\ \textcolor{blue}{Organization}', '\textcolor{blue}{duty-free sales}', '\textcolor{blue}{multiple-entry}\\ \textcolor{blue}{visas}', 'foreign tourists in Japan', 'Manryo Inc.\\ hostels', '\textcolor{blue}{Japanese economy}', '\textcolor{blue}{cultural}\\ \textcolor{blue}{attractions}', 'Chinese middle class', 'luxury\\ shopping in Japan', 'Sakaiminato tourism',\\ 'travel visas for Chinese tourists'{]}\end{tabular}} &
  \begin{tabular}[c]{@{}c@{}}{[}'\textcolor{blue}{Paul Manafort}', '\textcolor{blue}{Manafort Brothers Inc.}',\\ '{construction firm}', '\textcolor{blue}{New England politics}', 'New\\ Britain Connecticut', '\textcolor{blue}{special counsel’s}\\ \textcolor{blue}{investigation}', '\textcolor{blue}{Trump’s former campaign}\\ \textcolor{blue}{chairman}', '\textcolor{blue}{tax and bank fraud charges}',\\ '\textcolor{blue}{Ukrainian political interests}', '\textcolor{blue}{Washington}\\ \textcolor{blue}{lobbying firm}', 'Black Manafort and Stone',\\ '\textcolor{blue}{federal criminal and civil investigations}',\\ 'minority subcontractor allegations', 'perjury\\ charges', 'family legacy', 'community\\ foundation', 'Jessica Manafort', '\textcolor{blue}{Jess Bond}',\\ 'Rosy film'{]}\end{tabular} \\ \hline
\end{tabular}}
\caption{
Two example input documents from Inspec \cite{reips2002wextor}, \cite{moulin2002framework} and two from KPTimes \cite{japantimes2014}, \cite{nbcnews2019}, along with the corresponding keyphrase generations by various models, where blue represents present keyphrases and black represents absent keyphrases. These examples were chosen specifically to highlight the performance extremes across datasets: one demonstrating strong model performance and the other showcasing its limitations.} 
\label{tab:example2}
\end{table*}

\end{document}